\documentclass{article}

\usepackage[english]{babel}

\usepackage[letterpaper,top=2cm,bottom=2cm,left=3cm,right=3cm,marginparwidth=1.75cm]{geometry}

\usepackage{amsmath, amssymb, bm}
\usepackage{graphicx}
\usepackage[colorlinks=true, allcolors=blue]{hyperref}
\usepackage{rotating, threeparttable, multirow, multicol, makecell, booktabs}
\usepackage{array}
\usepackage{authblk}
\usepackage{indentfirst}
\usepackage{subcaption}
\usepackage{pgfplots}
\usepackage{cite}
\usepackage{float}

\definecolor{ntcol}{rgb}{0.0, 0.62, 0.24}

\usepackage{soul} 
\newif\ifhighlight
\highlightfalse  
\sethlcolor{yellow!25}
\newcommand{\modified}[1]{%
  \ifhighlight
    \hl{#1}%
  \else
    #1%
  \fi
}

\usepackage{tcolorbox}
\tcbuselibrary{skins, breakable}
\newenvironment{modifiedblock}
{%
  \ifhighlight
    \begin{tcolorbox}[
      colback=yellow!25,
      colframe=yellow!25,
      boxrule=0pt,
      arc=2pt,
      left=4pt,
      right=4pt,
      top=4pt,
      bottom=4pt,
      breakable
    ]
  \else
    \begingroup
  \fi
}
{%
  \ifhighlight
    \end{tcolorbox}
  \else
    \endgroup
  \fi
}

\usepackage[table]{xcolor}

\title{\textit{SuperWing}: a comprehensive transonic wing dataset for data-driven aerodynamic design}
\author[1,2]{Yunjia Yang}
\author[1]{Weishao Tang}
\author[1]{Mengxin Liu}
\author[2]{Nils Thuerey}
\author[1]{Yufei Zhang}
\author[1]{Haixin Chen\thanks{Corresponding author: chenhaixin@tsinghua.edu.cn}}
\affil[1]{School of Aerospace Engineering, Tsinghua University, Beijing, China}
\affil[2]{School of Computation, Information and Technology, Technical University of Munich, Germany}

\newcolumntype{M}[1]{>{\centering\arraybackslash}m{#1}} 

\begin{document}

\maketitle

\begin{abstract}
    Machine-learning surrogate models have shown promise in accelerating aerodynamic design, yet progress toward generalizable predictors for three-dimensional wings has been limited by the scarcity and restricted diversity of existing datasets. Here, we present \textit{SuperWing}, a comprehensive open dataset of transonic swept-wing aerodynamics comprising 4,239 parameterized wing geometries and 28,856 Reynolds-averaged Navier-Stokes flow field solutions. The wing shapes in the dataset are generated using a simplified yet expressive geometry parameterization that incorporates spanwise variations in airfoil shape, twist, and dihedral, allowing for an enhanced diversity without relying on perturbations of a baseline wing. All shapes are simulated under a broad range of Mach numbers and angles of attack covering the typical flight envelope. To demonstrate the dataset’s utility, we benchmark two state-of-the-art Transformers that accurately predict surface flow and achieve a 2.5 drag-count error on held-out samples. Models pretrained on \textit{SuperWing} further exhibit strong zero-shot generalization to complex benchmark wings such as DLR-F6 and NASA CRM, underscoring the dataset’s diversity and potential for practical usage.
\end{abstract}

\section{Background \& Summary}


Machine learning (ML) techniques have emerged over the past decade as a powerful tool to reshape aerodynamic design optimization. By learning directly from data to predict flow fields given shape and operating condition, ML models can serve as surrogate models~\cite{tao2019application,thuerey2020dfp} in iterative shape-optimization loops, significantly reducing computational cost and accelerating the design process~\cite{li_dbo_2019, li_dbo_2021, chen2021numerical}. They can be also used to provide designers with near-instant responses of flow field and performance, enabling interactive optimization and generative design. 

\renewcommand{\arraystretch}{1.5}
\begin{sidewaystable}
\centering
\begin{threeparttable}
\caption{\label{tab:existing}Summery of existing wing dataset and models}
\begin{tabular}{c|cM{5cm}cM{2.5cm}M{3cm}}
\hline\hline
Citation & \# of samples & Input parameters \tnote{a,b,c} & CFD solver  & Model \tnote{d} & Baseline wing \tnote{e} \\\hline
Li et al. \cite{li_dbo_2021} & 183,075 & $ Ma, \alpha$, wing modes (40)  & ADflow & MLP & NASA CRM \\
Castellanos et al. \cite{castellanos_wing_2022} & 531 & $Ma, \alpha$ & TAU & POD, RF, MLP & Airbus XRF1 \\
Lyu et al. \cite{lyu_multi-fidelity_2023} & 2,000 & $Re, V_\infty, \alpha$ & OpenFOAM & MF-FNO & falcon\\
Immordino et al. \cite{immordino_wing_2024, immordino_predicting_2024} & $70 \times 3$ & $Ma, \alpha$ & SU2 & POD, CNN, GNN, BN & BSCW, ONERA M6, NASA CRM \\
Hines et al. \cite{hines_gnn_wing_2023} & 157 & $Ma, \alpha, \delta_\text{ail}$(4) & TAU & GNN & NASA CRM\\
Hu et al. \cite{hu_wing1_2023} & 200 & CSTs (36) & PHengLEI & CNN & ONERA M6\\
Kim et al. \cite{kim_wing2_2023} & 735 & $\alpha, AR$, L.E. wavy para. & Star CCM+ & CNN & ISW\\
Li et al. \cite{li_transfer_2023} & 10,000 & $ Ma, \alpha$, CSTs (14), planform (7) & BLWF & MLP & \textbf{None}\\
Catalani et al. \cite{catalani_neural_2024} & 8,640 (120 shapes) & $Ma, \alpha, b, t, \Gamma, \delta_\text{ail}$ & BLWF & INR & Airbus XRF1\\
Lei et al. \cite{lei_prediction_2024} & 363 & $Ma, \alpha, \Lambda$ & / & RBFN-CAE &  BQM-34 \\
Yang et al. \cite{yang_transferable_2024} & 846 + 164 & $Ma, \alpha$ planform (6 or 8) & CFL3D & UNet & AIAA DPW-W1\\
Yang et al. \cite{yang_rapid_2025} & 1,842 & $Ma, \alpha$ CSTs (14), planform (7) & CFL3D & UNet & \textbf{None}\\
Zuo et al. \cite{zuo_flow3dnet_2025} & 800 & CSTs & PHengLEI & UNet & ONERA M6 \\
Hasan et al. \cite{hasan_wing_bspline_spanwise_crm_2025} & 160 & planform (10) & Ansys fluent & LGBM & NASA CRM \\\hline
\textbf{Present} & \textbf{28,856 (4,239 shapes)} & $Ma, \alpha$, \textbf{CSTs (20), planform (5), spanwise variation (13)}  & \textbf{ADflow} & \textbf{Transformer} & \textbf{None}\\\hline\hline
\end{tabular}
\begin{tablenotes}
\footnotesize
\item[a] Abbreviations: CST (Class shape transformation), L.E. (Leading edge)
\item[b] Symbols: $Ma$ (Mach number), $\alpha$ (Angle of attack), $\delta_\text{ail}$ (deflection of ailerons), $AR$ (Aspect ratio), $b$ (span), $t$ (thickness), $\Gamma$ (dihedral angle), $\Gamma$ (Sweep angle)
\item[c] the numbers in the parentheses are parameters with more than one dimensions
\item[d] Abbreviations: MLP (Multilayer perceptrons), CNN (convolutional neural network), INR (Implicit neural representation), LGBM (light gradient boosting machine), GNN (graph neural network), POD (Proper Orthogonal Decomposition), BN (Bayesian network)
\item[e] Abbreviations: CRM (Common research model), ISW (infinite swept wing), BSCW (Benchmark Super Critical Wing)
\end{tablenotes}
\end{threeparttable}
\end{sidewaystable}

The above applications rely on foundation models that are generalizable across diverse shapes and operating conditions. In recent years, researchers have successfully trained such models for two-dimensional aerodynamic components, leading to notable progress in design optimization \cite{yang_fast_2024, li_dbo_2019, bouhlel_scalable_2020, renganathan_enhanced_2021, tian_dbo_pressureguide_2024}. However, developing a foundation model capable of predicting flow fields and aerodynamic performance for three-dimensional configurations such as wings remains a significant challenge. The primary limitation is the lack of sufficiently diverse datasets that capture flow fields across a wide range of wing geometries. Existing datasets predominantly focus on turbulent flows around a few fixed configurations \cite{data_mcconkey2021curated, data_zhang2025high}, and those that include multiple wing shapes still suffer from limited geometric diversity. We summarized wing flow field datasets in Table \ref{tab:existing}, where some studies employ fixed geometries while varying only the flow conditions \cite{castellanos_wing_2022, lyu_multi-fidelity_2023, immordino_wing_2024, immordino_predicting_2024}, while others perturb a single baseline wing to generate training samples \cite{li_dbo_2021, hines_gnn_wing_2023, hu_wing1_2023, kim_wing2_2023, catalani_neural_2024, lei_prediction_2024, yang_transferable_2024, zuo_flow3dnet_2025, hasan_wing_bspline_spanwise_crm_2025}. Such approaches restrict the model’s general applicability, since a foundation model is expected to operate across a wide variety of wing geometries. In our previous work \cite{li_transfer_2023, yang_rapid_2025}, we attempted to build datasets independent of a baseline wing, sampling shapes from a variety of sectional airfoils and planform parameters. While this represented a step forward, the resulting configurations still lacked the complexity of realistic designs.

Another bottleneck lies in the limitations of conventional ML architectures when applied to large-scale aerodynamic datasets. Recent advances in Transformer-based neural networks have demonstrated strong scalability and the ability to capture complex correlations \cite{wu_transolver_2024, herde_poseidon_2024, xu_self-supervised_2024, holzschuh_pde-transformer_2025, zhou_unisolver_2025, luo_mmet_2025}. However, most existing studies focus on relatively simple geometries, and their effectiveness on realistic engineering problems remains underexplored.

\modified{To address these challenges, this study contributes a comprehensive dataset for clean transonic wings with approximately 30,000 flow-field samples that aim to benefit both the machine learning and the aerospace communities. Guided by a survey of representative industrial wing designs, including two benchmarks and two real-world configurations, we propose a simplified yet expressive geometric parameterization scheme for the wing shapes to generate the dataset. Unlike baseline-perturbation approaches, our method constructs geometries from scratch using design rules grounded in realistic transonic wing configurations. This ensures that the dataset is both diverse and engineering-practical. }

\begin{modifiedblock}
\begin{itemize}
    \item \textbf{For the machine-learning community}, \textit{SuperWing} provides a large-scale and sufficiently complex benchmark for learning three-dimensional aerodynamics. The dataset includes both surface and volumetric point-cloud flow data with relatively high resolution. The surface-flow data is also provided in a structured-mesh format, enabling benchmarking for architectures that require structured inputs and outputs. On this basis, we benchmark two state-of-the-art Transformer architectures, Vision Transformer (ViT) \cite{dosovitskiy_vit_2021} and Transolver \cite{wu_transolver_2024}, demonstrating that the dataset supports a fair comparison of advanced ML models on a realistic three-dimensional aerodynamic prediction problem.

    \item \textbf{For the aerospace community}, the scale and diversity of \textit{SuperWing} provide an opportunity to pre-train \textit{large-scale surrogate models} and adapt them to downstream aerodynamic design tasks, such as interactive design \cite{bouhlel_scalable_2020} and data-based optimization \cite{li_dbo_2019}. Although \textit{SuperWing} is generated using a simplified geometric representation, the validation results show that models trained on it can exhibit promising zero-shot generalization to more complex benchmark wings, such as CRM \cite{vassberg_development_2008} and DLR-F6 \cite{vassberg_f6_2005}. This indicates that the model has learned general transonic aerodynamic principles from the large-scale data. More broadly beyond the clean wings, these principles are also held. A model pre-trained on \textit{SuperWing} could serve as a base model, and be fine-tuned with fewer additional samples for more complex use cases, such as wing-body or wing-nacelle configurations. Since generating large-scale datasets for such complex configurations incurs substantially higher computational costs, pre-training on \textit{SuperWing} can help reduce the computational burden for downstream applications, enabling practical data-driven aerodynamic design.

\end{itemize}
\end{modifiedblock}

The dataset is released at \href{https://huggingface.co/datasets/yunplus/SuperWing}{https://huggingface.co/datasets/yunplus/SuperWing}.

\section{Methods}\label{sec:method}

The \textit{SuperWing} dataset does not have a single baseline wing shape, but instead focuses on a specific wing planform shape: the “Yehudi break” \cite{vassberg_development_2008}, or “kink” shape. 
To make the wings in the dataset resemble realistic wings, they are built by first selecting a sectional airfoil, then sampling the planform shape and the parameters (including the dihedral angle, twist angle, thickness, and camber) that vary along the span. In this section, we first introduce the method for building a wing, summarize several representative configurations commonly used in the aircraft industry, and finally present the parameter distributions of the dataset.

\subsection{Sampling of wing shapes}

\subsubsection{Geometric parameters of wings}

A wing is constructed by stretching the sectional airfoils along the spanwise direction. In principle, the airfoils at each spanwise station could differ entirely, leading to a highly high-dimensional parameter space. To balance diversity and simplicity, we identify common spanwise variation patterns and incorporate them into the wing geometry description.

Specifically, each wing is generated from a single baseline airfoil, parameterized using a 9th-order Class-Shape Transformation (CST) method \cite{kulfan_universal_2008}. The upper and lower surfaces are represented independently. 

\begin{equation}
    y_\text{u,b}(x) = C(x) \cdot \sum_{i=0}^{9} u_i \cdot \Phi_i(x),\quad     y_\text{l,b}(x) = C(x) \cdot  \sum_{i=0}^{9} l_i \cdot \Phi_i(x)
\end{equation}
\modified{where $C(x) = x^{0.5}(1-x)^{1.0}$ is the class function, and where $\phi_i(x)=\binom{n}{i}x^i(1-x)^{n-i}$ denotes the $i$-th Bernstein basis polynomial of degree $n$.}

The baseline airfoil can also be described with its normalized thickness and camber line:

\begin{equation}
    \tilde t_\text{b}(x) = \frac{y_\text{u,b}(x) - y_\text{l,b}(x)}{t_{\max,\text{b}}}, \quad \tilde \delta_\text{b}(x) = \frac{0.5 \left(y_\text{u,b}(x) + y_\text{l,b}(x)\right)}{\delta_{\max,\text{b}}}
\end{equation}
where $t_{\max,\text{b}} = \max_x\left(y_\text{u,b}(x) - y_\text{l,b}(x)\right)$ and $\delta_{\max,\text{b}} = 0.5 \max_x\left(y_\text{u,b}(x) + y_\text{l,b}(x)\right)$.

Since thickness and camber are essential to determining an airfoil’s aerodynamic performance, the spanwise distribution of maximum thickness and maximum camber is prescribed as functions of the spanwise station $\eta$, i.e., $t_{\max}(\eta)$ and $\delta_{\max}(\eta)$. This allows the construction of airfoil sections across the span based on the baseline airfoil and these distribution functions:

\begin{equation}
    t(x,\eta) = t_{\max}(\eta) \cdot \tilde t_\text{b}(x), \quad \delta(x,\eta) = \delta_{\max}(\eta) \cdot \tilde \delta_\text{b}(x)
\end{equation}

\begin{figure}[htbp]
    \centering
    \includegraphics[width=0.9\linewidth]{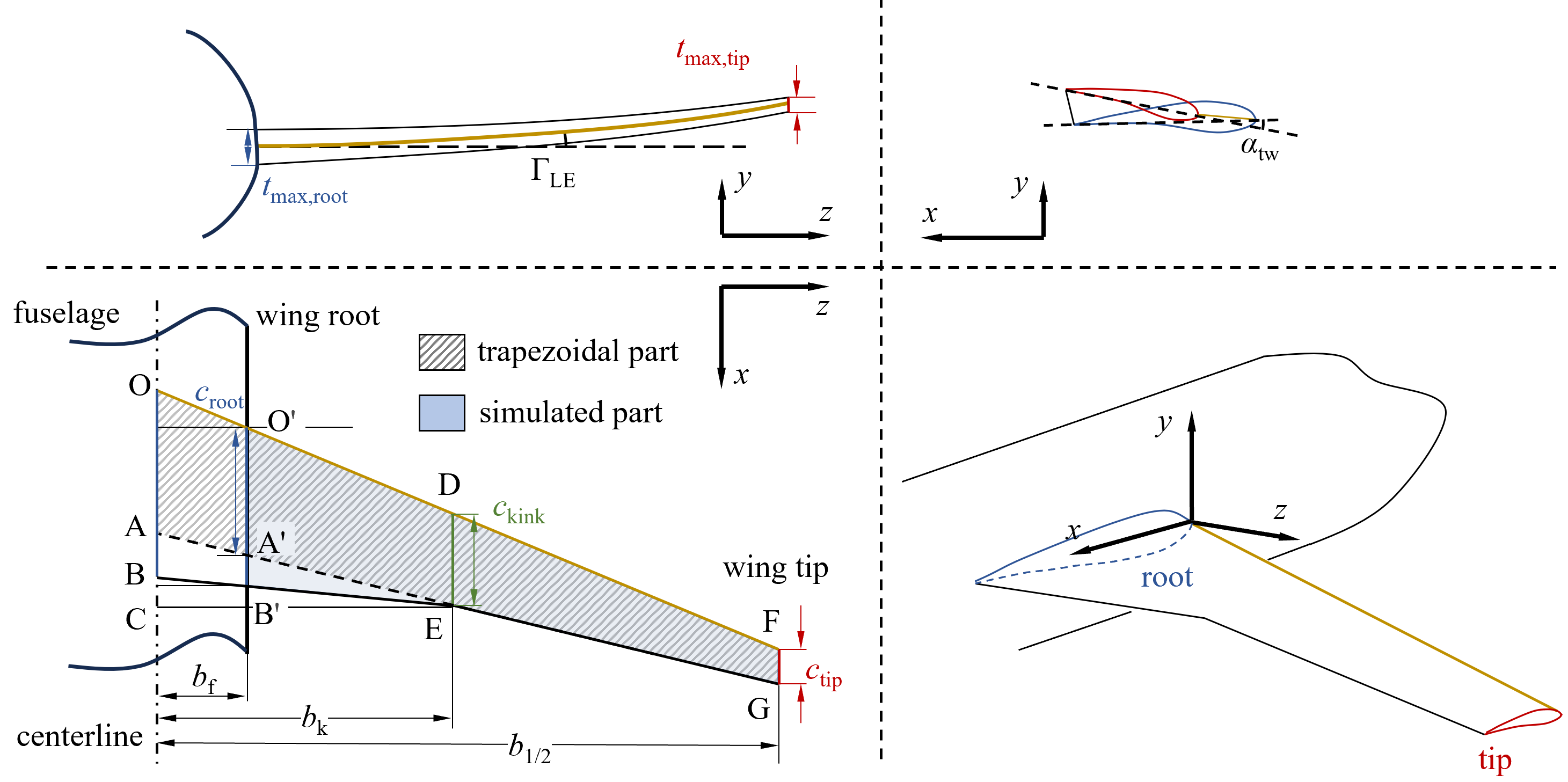}
    \caption{Three-view diagram of a typical kink wing}
    \label{fig:three-view}
\end{figure}

The parameters that define the planform shape are illustrated in the three-view diagram in Fig. \ref{fig:three-view}. From the top view, the wing can be seen as a combination of a trapezoidal part ($OAGF$, marked by right slash lines) and an extra surface part between the root and the kink ($\triangle ABE$). The trapezoidal part shape is determined with aspect ratio $AR = \frac{2b_{1/2}^2}{S_\text{trap}} = \frac{4b_{1/2}}{(\overline{FG}+\overline{OA})}$, taper ratio $TR = \frac{\overline{FG}}{\overline{OA}}$, and the leading-edge sweep angle $\Lambda_\text{LE}$. The extra surface can be determined with the kink location $\eta_\text{k}=\frac{b_\mathrm{k}}{b_{1/2}}$ and root adjustment ratio $\kappa = \frac{\overline{AB}}{\overline{AC}}$. 

In the present study, the fuselage location is fixed at 10\% span, and only the exposed part of the wing to the air is simulated. The simulation part corresponds to the light blue area ($O'B'EGF$). The wing's reference area is the simulation part's projection area. 

Dihedral angles $\Gamma_\text{LE}(\eta)$ and twist angles $\alpha_\text{tw}(\eta)$ control the $y$-positions and the rotation angles of every section airfoil. Modern wings always vary along the spanwise direction to achieve the best aerodynamic performance.

\subsubsection{Summary of the parameters of several typical wings}

Based on the definitions of wing parameters in the last section, four wing benchmarks and real aircraft wings are studied, and their shape parameters are listed in Table \ref{tab:wings}. The benchmark models include the DLR-F6 model from the German Aerospace Center \cite{vassberg_f6_2005}, and NASA's Common Research Model (CRM)\cite{vassberg_development_2008}. The real-world wings are from the Airbus A320 \cite{orlita_a320_2017} and Boeing 787 (obtained from the free sample of the software Piano: \href{https://www.lissys.uk/samp1/b787.html}{https://www.lissys.uk/samp1/b787.html}). 

\begin{table}[htbp]
\small
    \centering
    \begin{threeparttable}
    \caption{\label{tab:wings}Geometric parameters of typical wings}
    \begin{tabular}{c|ccccc}
        \hline\hline
        Parameter & DLR-F6 & CRM & A320 & B787 \\\hline
        Cruise Mach number & 0.75 & 0.85 & 0.775 & 0.85 \\
        max. relative thickness at centerline & 0.1629 & 0.1542 & 0.1394 & 0.1449 \\
        ratio of $t_{\max}$ at kink and centerline & 0.7316 & 0.6822 & — & 0.6472 \\
        ratio of $t_{\max}$ at tip and centerline & 0.7306 & 0.6161 & 0.7166 & 0.6056 \\
        taper ratio \tnote{a} & 0.380 & 0.275 & 0.330 & 0.180 \\
        L.E. swept angle (deg) & 25.15 & 37.16 & 25.00 & 32.20 \\
        L.E. dihedral angle at tip (deg) & 5.2 & 4.4 & 4.4 & 6.0 \\
        aspect ratio \tnote{b}  & 9.28 & 8.38 & 8.79 & 9.20 \\
        kink location (\%)  & 40.1 & 37.0 & 39.2 & 37.4 \\
        root adjustment  & 1.00 & 0.67 & 1.00 & 0.88 \\
        twist angle at tip  & 6.14 & 10.47 & 3.82 & — \\\hline\hline
    \end{tabular}
    \begin{tablenotes}
    \footnotesize
    \item[a] based on trapezoidal wing
    \item[b] based on total projection area
    \end{tablenotes}
    \end{threeparttable}
\end{table}

\begin{figure}[ht]
    \centering
    \begin{subfigure}{0.4253\textwidth}
        \centering
        \includegraphics[width=\linewidth]{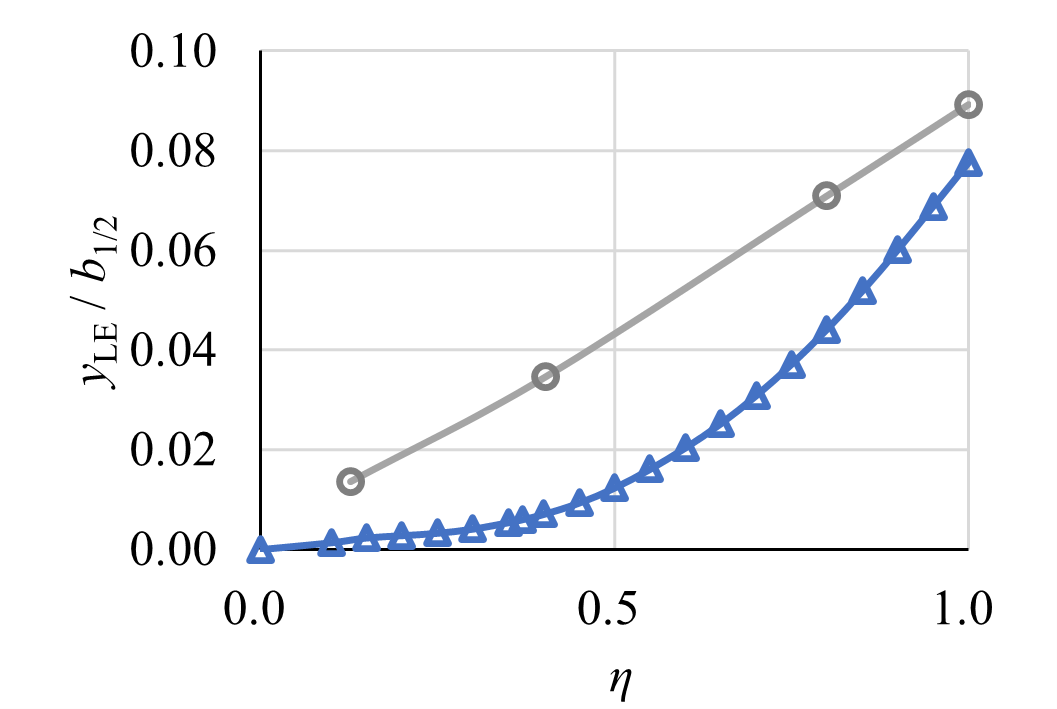}
        \caption{local dihedral ratio}
        \label{fig:sub-a}
    \end{subfigure}
    \hfill
    \begin{subfigure}{0.5387\textwidth}
        \centering
        \includegraphics[width=\linewidth]{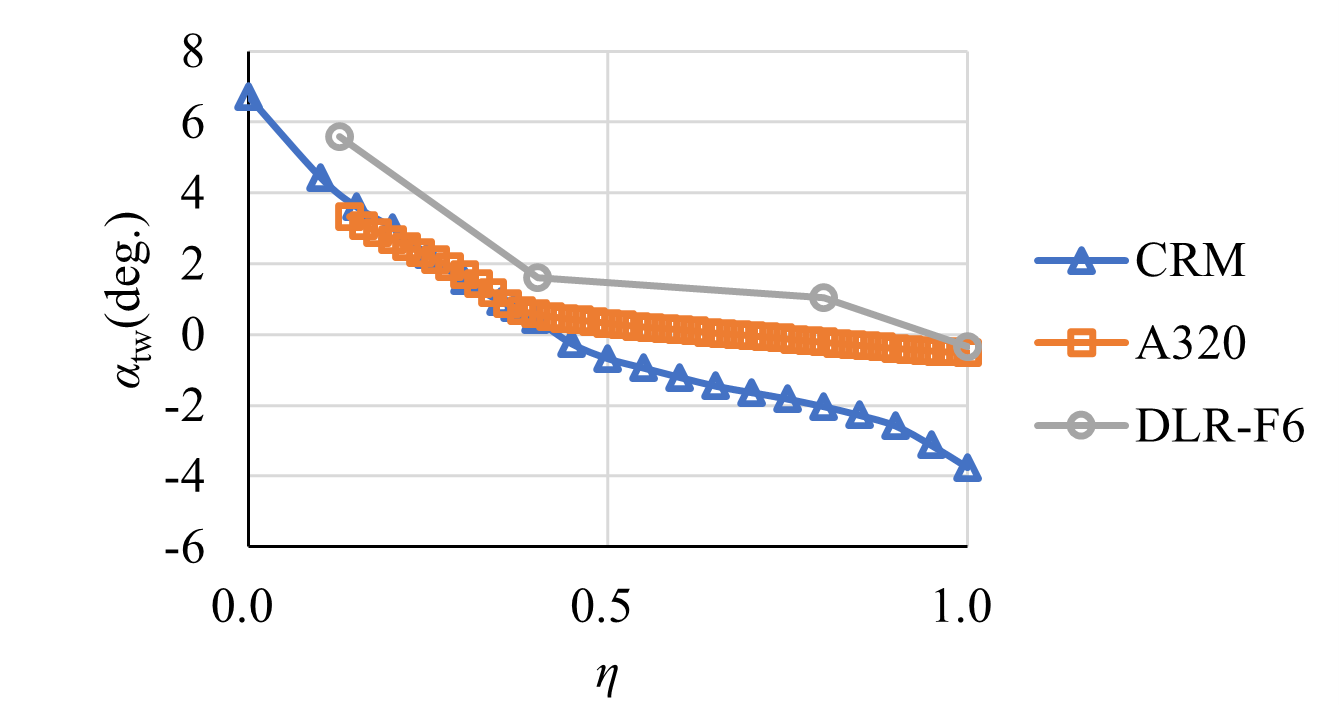}
        \caption{local twist angle}
        \label{fig:sub-b}
    \end{subfigure}
        
    \begin{subfigure}{0.4253\textwidth}
        \centering
        \includegraphics[width=\linewidth]{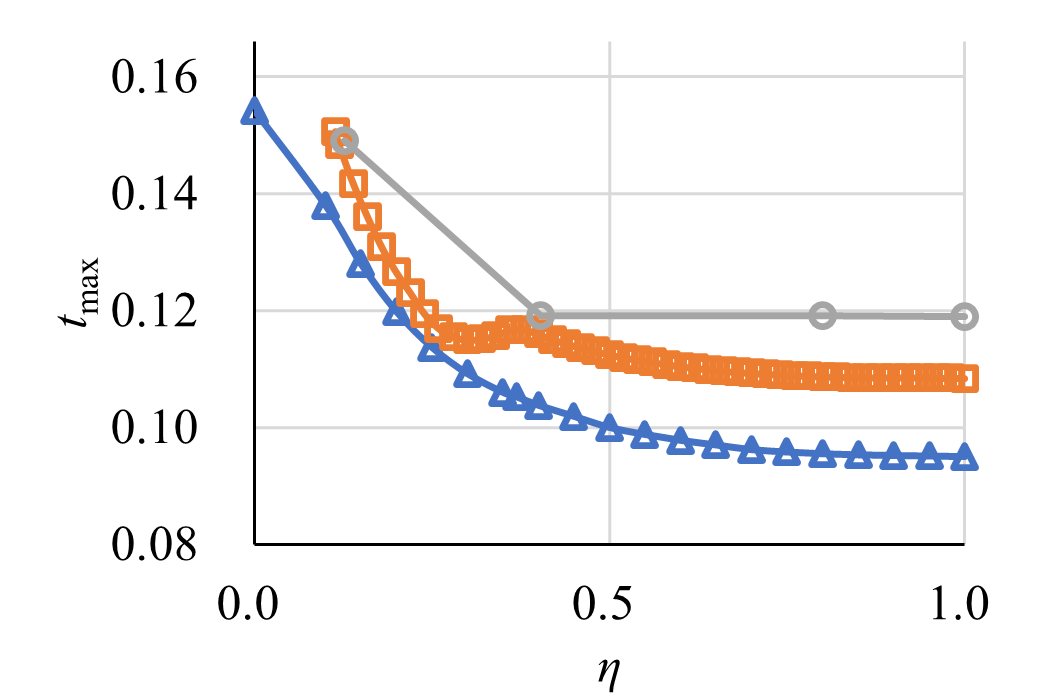}
        \caption{local maximum relative thickness}
        \label{fig:sub-c}
    \end{subfigure}
    \hfill
    \begin{subfigure}{0.5387\textwidth}
        \centering
        \includegraphics[width=\linewidth]{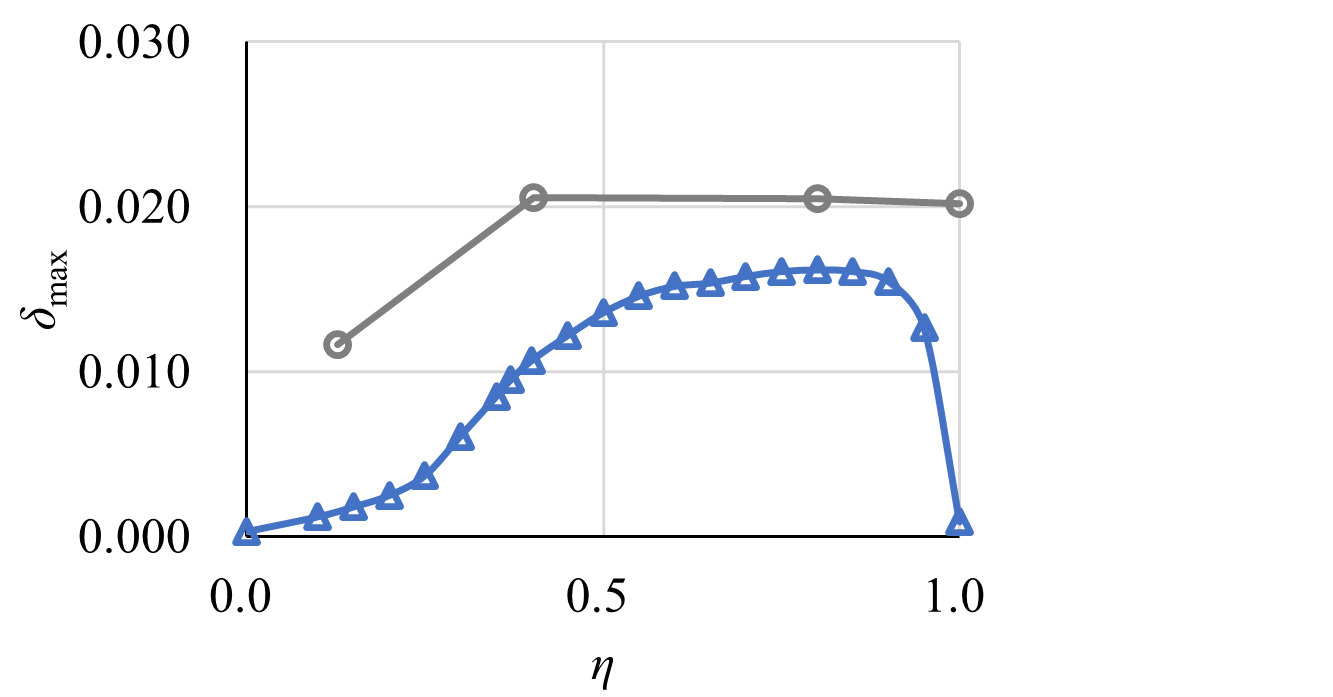}
        \caption{local maximum relative camber}
        \label{fig:sub-d}
    \end{subfigure}
    \caption{Spanwise distribution of parameters for typical kinked wings}
    \label{fig:etawings}
\end{figure}

As mentioned above, the dihedral angle, twist angle, maximum thickness, and camber along the span are varied. Figure \ref{fig:etawings} summarizes these spanwise distributed parameters for the wings.

\subsubsection{Sampling of the parameters}

By analyzing the typical wing configurations, we determine the parameter ranges and use them as the basis for sampling to establish our dataset.

\begin{figure}[htbp]
    \centering
    \includegraphics[width=0.8\linewidth]{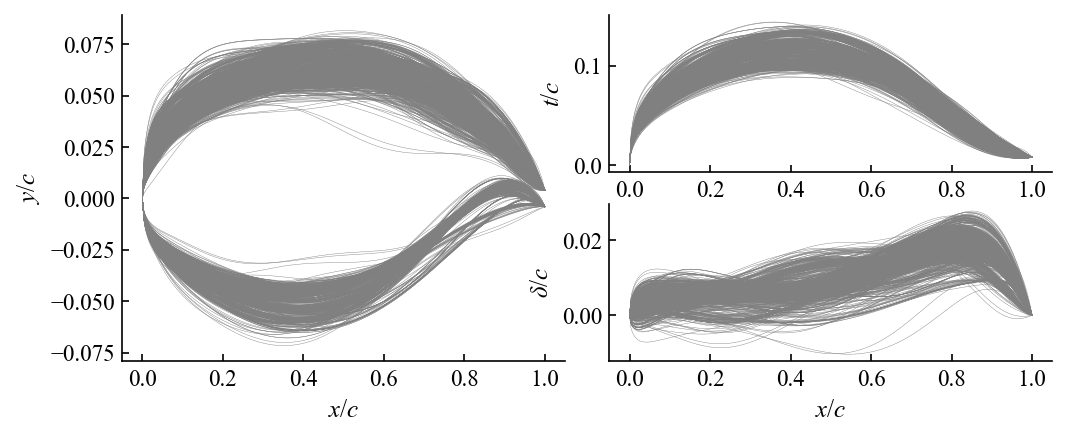}
    \caption{Airfoil shapes and their thickness and camber distributions}
    \label{fig:airfoils}
\end{figure}

\paragraph{Baseline airfoil} We utilize an existing database \cite{yang_rapid_2025} to generate the CST coefficients for the baseline airfoils. They are sampled using the Output Space Sampling (OSS) method \cite{li_oss_2022}, which aims to produce geometric variations that exhibit diverse and representative pressure distribution patterns. \modified{Their airfoil CFD solutions were checked during sampling, and cases with nonphysical flow fields were removed.} A total of 1,516 sets of CST coefficients are obtained. Figure \ref{fig:airfoils} presents the corresponding normalized thickness $\tilde t(x)$ and camber lines $\tilde \delta (x)$ of the sampled airfoils.

\paragraph{Planform parameters} The planform parameters are randomly sampled from ranges shown in Table \ref{tab:ranges0}, which cover the typical wings. 

\begin{table}[htbp]
    \small
    \centering
    \caption{Sampling ranges of the wing planform parameters}
    \begin{tabular}{cccc}
        \hline\hline
        Parameter & Symbol & Lower range & Upper range \\ \hline
        sweep angle & $\Lambda_\mathrm{LE}$ & $25^\circ$ & $40^\circ$ \\ 
        aspect ratio (trap.) & $AR$ & 8 & 11 \\
        taper ratio (trap.) & $TR$ & 0.15 & 0.40 \\
        kink location & $\eta_k$ & 36\% & 42\% \\
        root adjustment & $\kappa_\mathrm{root}$ & 10\% & 110\% \\
        \hline\hline
    \end{tabular}
    \label{tab:ranges0}
\end{table}


\paragraph{Spanwise-variation parameters} The spanwise variations of dihedral angle, twist angle, maximum thickness, and maximum camber are determined using cubic spline interpolation based on five spanwise control points (CPs). A similar methodology has been applied to optimizing wing shapes\cite{hasan_wing_bspline_spanwise_crm_2025}. 

\begin{figure}[ht]
    \centering
    \includegraphics[width=0.5\linewidth]{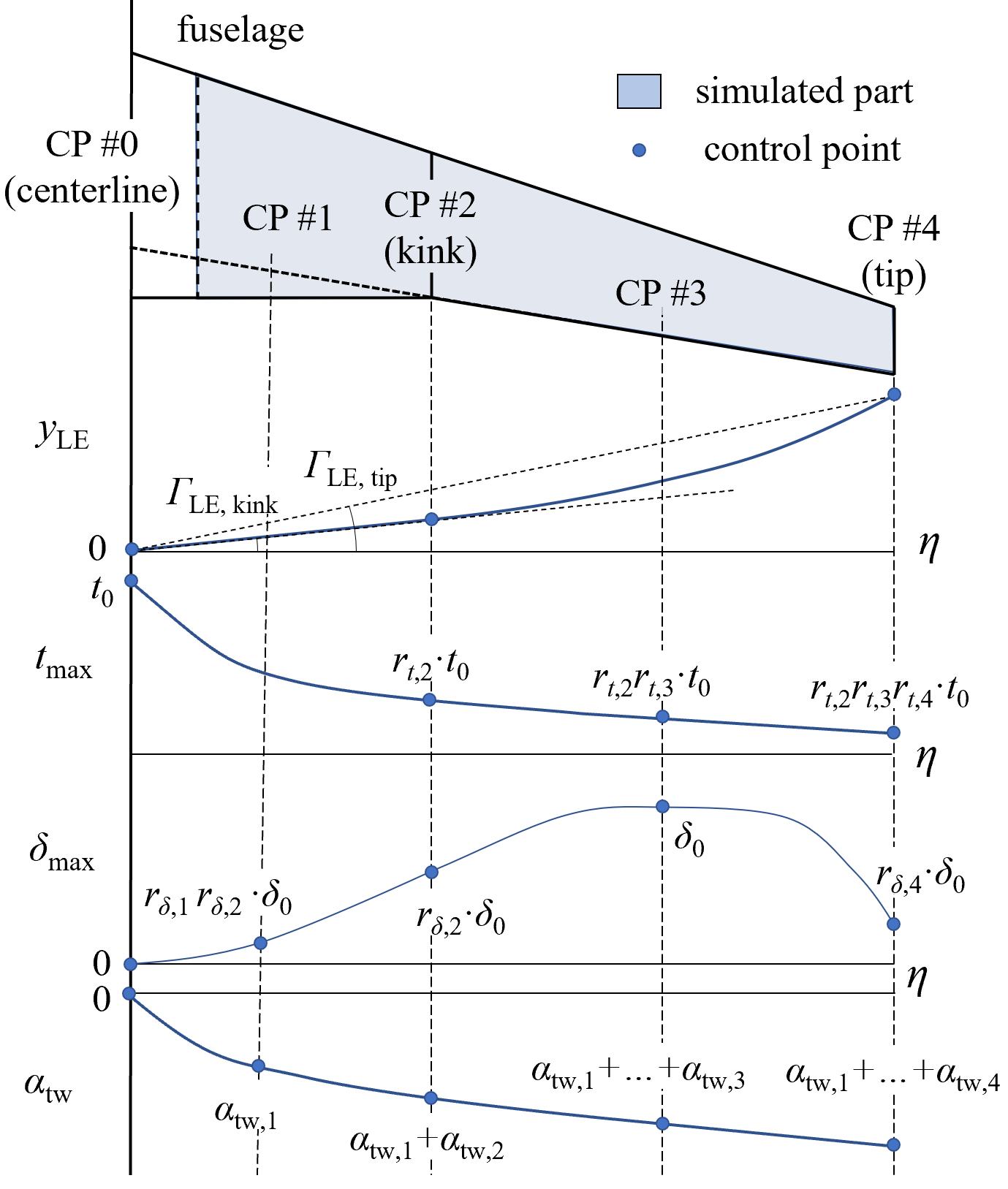}
    \caption{Spanwise control points and the variables to generate distributed wing parameters}
    \label{fig:spline}
\end{figure}

As illustrated in Figure \ref{fig:spline}, the CPs 0 to 4 extend from the centerline to the tip. Specifically, CPs \#0, \#2, and \#4 are at the root, kink, and tip, while CPs \#1 and \#3 are located in the middle between adjacent points.

\begin{itemize}
    \item \textbf{Dihedral angle:} The dihedral angle is defined by CPs \#2 and \#4. The leading-edge $y$-coordinates are initialized to zero at the centerline and increase linearly toward the kink, with a slope defined by the ratio $\frac{y_\text{LE}(\eta)}{b(\eta)} = \tan(\Gamma_\text{LE,kink})$. The leading edge $y$-coordinate of the tip is determined by $\frac{y_\text{LE,tip}}{b_{1/2}} = \tan(\Gamma_\text{LE,tip})$. The intermediate segment between kink and tip is generated using a cubic spline with a slope matched to the linear segment to ensure continuity.
    \item \textbf{Airfoil thickness and camber:} The distribution of maximum thickness $t_{\max}(\eta)$ is defined using CPs \#0, \#2, \#3, and \#4. The thickness at CP \#0, along with the ratios of thickness values at the remaining control points relative to CP \#0, denoted $t_0$, $r_{t,2}$, $r_{t,3}$, and $r_{t,4}$, are used to construct the spline. Similarly, maximum camber distribution $\delta_{\max}(\eta)$ is defined from CPs \#0 to \#4, with values fixed at zero for CP \#0 and at $\delta_0$ for CP \#3. The maximum cambers at CPs \#1, \#2, and \#4 are expressed as ratios relative to $\delta_{0}$, denoted $r_{\delta,1}$, $r_{\delta,2}$, and $r_{\delta,4}$. 
    \item \textbf{Twist angle:} Twist angles are also specified at all five CPs, with the twist at CP \#0 set to zero. The angles at outer control points are defined as incremental deviations from the root. 

\end{itemize}

The above method can describe the spanwise-distributed parameters with several coefficients. They are also randomly sampled from the ranges inferred from the typical wings, which are summarized in Table \ref{tab:ranges}. 

\begin{table}[htbp]
\small
    \centering
    \caption{Sampling ranges of spanwise-distributed parameters}
    \begin{tabular}{ccccc}
        \hline\hline
        \multicolumn{2}{c}{Parameter} & Symbol & Lower range & Upper range \\ \hline
        \multirow{2}{*}{dihedral angle} & tip & $\Gamma_\mathrm{LE,tip}$ & $4^\circ$ & $6^\circ$ \\
        & kink & $\Gamma_\mathrm{LE,kink}$ & $0.5^\circ$ & $\Gamma_\mathrm{LE,tip}$ \\
         \multicolumn{2}{c}{root max. relative thickness} & $(t/c)_\mathrm{root}$ & 0.14 & 0.17 \\
        \multirow{3}{*}{thickness ratio} & CP \#2 & $r_{t,2}$ & 0.60 & 0.70 \\
        & CP \#3 & $r_{t,3}$ & 0.90 & 0.98 \\
        & CP \#4 & $r_{t,4}$ & 0.92 & 1.00 \\
        \multirow{3}{*}{camber ratio} & CP \#1 & $r_{\delta,1}$ & 0.3 & 0.8 \\
        & CP \#2 & $r_{\delta,2}$ & 0.5 & 1.0 \\
        & CP \#4 & $r_{\delta,4}$ & 0.0 & 0.8 \\
        \multirow{3}{*}{twist angle} & CP \#1 & $\alpha_{\mathrm{tw},1}$ & $-4^{\circ}$ & $-2^{\circ}$ \\
        & CP \#2 & $\alpha_{\mathrm{tw},2}$ & $-4^\circ$ & $-2^\circ$ \\
        & CP \#3 & $\alpha_{\mathrm{tw},3}$ & $-3^\circ$ & $-1^\circ$ \\
        & CP \#4 & $\alpha_{\mathrm{tw},4}$ & $-3^\circ$ & $-1^\circ$ \\
        \hline\hline
    \end{tabular}
    \label{tab:ranges}
\end{table}

\begin{figure}[ht]
    \centering
    \begin{subfigure}{0.65\textwidth}
        \centering
        \includegraphics[width=\linewidth]{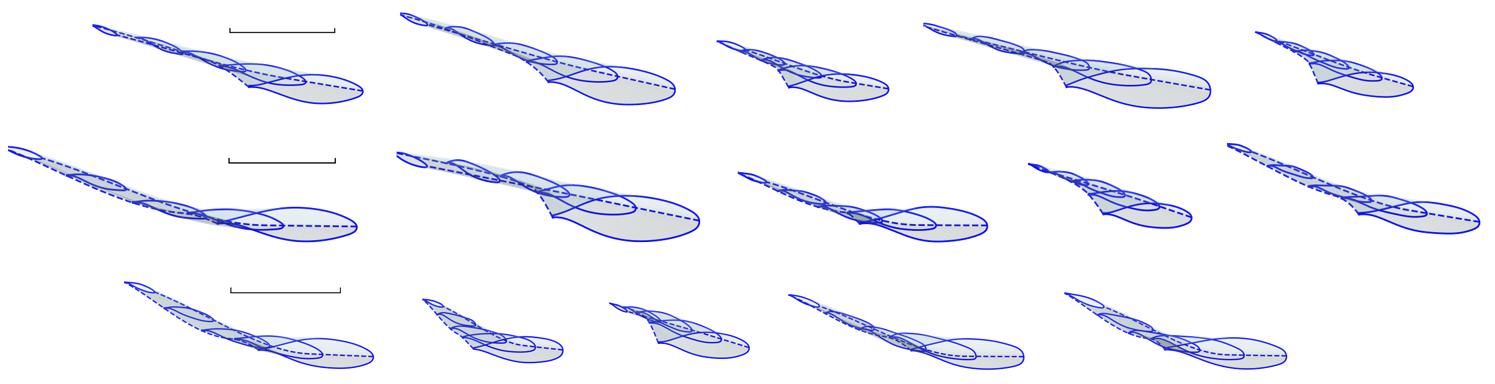}
        \caption{Airfoil stack plot}
        \label{fig:stacks}
    \end{subfigure}
    \hfill
    \begin{subfigure}{0.34\textwidth}
        \centering
        \includegraphics[width=\linewidth]{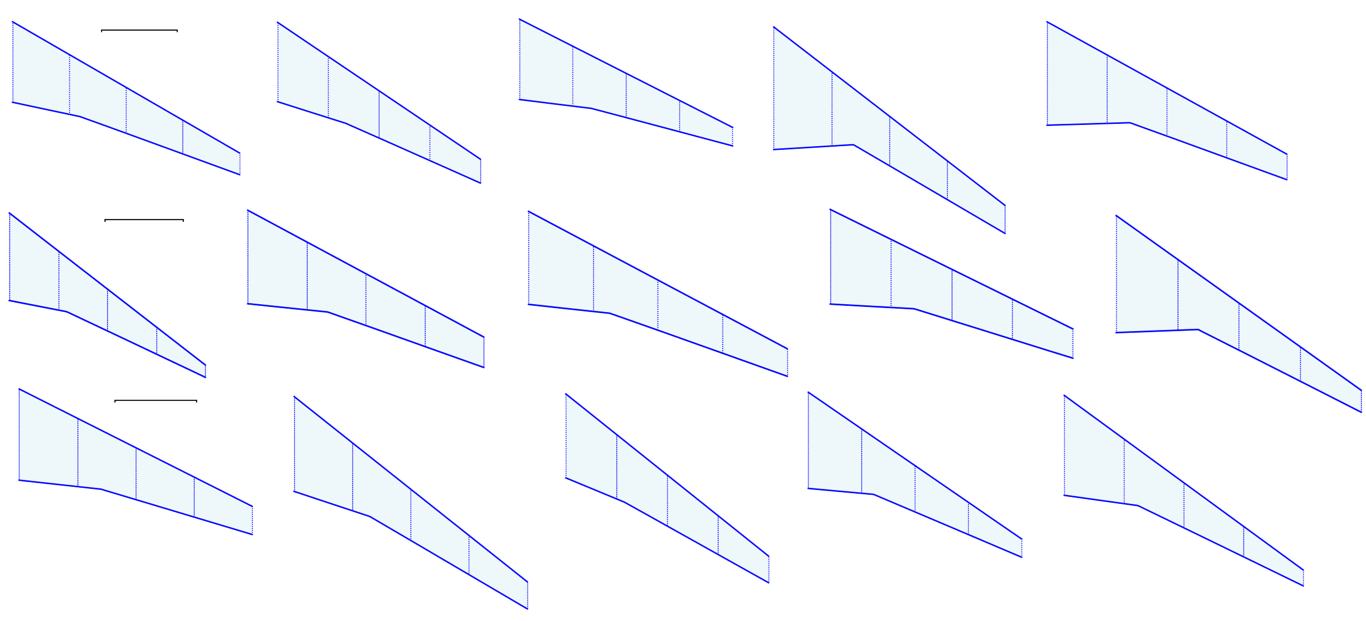}
        \caption{Top views}
        \label{fig:tops}
    \end{subfigure}
    \caption{Visualization of several wing shapes in the dataset}
    \label{fig:wings}
\end{figure}

In conclusion, each wing geometry is uniquely defined by 
the CST coefficients of the baseline airfoil, and 18 additional parameters. For each set of CST coefficients, three independent sets of spanwise parameters are sampled, resulting in about 4500 distinct wing geometries. Figure \ref{fig:wings} shows a collection of several randomly selected wing shapes in the dataset. 

It's worth mentioning that, although each baseline sectional airfoil in the proposed dataset is associated with only three sets of spanwise and planform parameters, their combinatorial variations provide broad coverage of the overall wing shape design space. As supported by our prior experience \cite{yang_rapid_2025} and the experimental evidence presented later in this work, machine-learning models are able to learn the underlying physical relationships that generalize these sectional geometries to other global parameter settings. 

\subsection{Sampling of wing operating conditions}

For each wing geometry, eight operating conditions \modified{are independently sampled from uniform distributions. Given the large sample size, we use random sampling to allow the dataset to be incrementally extended without redesigning the entire sampling scheme.} The freestream Mach number $Ma$ ranges from 0.75 to 0.90, and the angle of attack $\alpha$ varies between $2^\circ$ and $12^\circ$. The Reynolds number and freestream temperature are fixed at 20 million and 300 K, respectively, for all simulations.

\subsection{Simulation of the wing flow fields}

Reynolds-Averaged Navier-Stokes (RANS) simulations are performed using the open-source CFD solver suite developed by the \verb|MDOLab| (\href{https://mdolab.engin.umich.edu/software}{https://mdolab.engin.umich.edu/software}).

\subsubsection{CFD setting}

The flow field around each wing is computed using \verb|ADflow| \cite{mader_adflow_2020}. \modified{The governing equations are RANS using the Spalart-Allmaras turbulence model in a fully turbulent setting. The root boundary is treated as the symmetry plane, and the wing surface is treated as the adiabatic wall. A “3w” multigrid strategy is adopted to accelerate convergence: 500 solver cycles are applied on the coarser multigrid levels, followed by up to 3,000 cycles on the original fine mesh. Initially, the approximate Newton-Krylov (ANK) solver is used to solve the linear systems; once the total residual drops below $1\times10^{-8}$, the solver switches to the Newton-Krylov (NK) method. Convergence is declared when the residual falls below $1\times10^{-10}$. All other solver settings follow the default configuration, and we provide the simulation script on GitHub} (\href{https://github.com/tum-pbs/AeroTransformer}{https://github.com/tum-pbs/AeroTransformer}).

\subsubsection{Mesh generation} \label{sec:mesh}

\begin{figure}[htbp]
    \centering
    \includegraphics[width=0.5\linewidth]{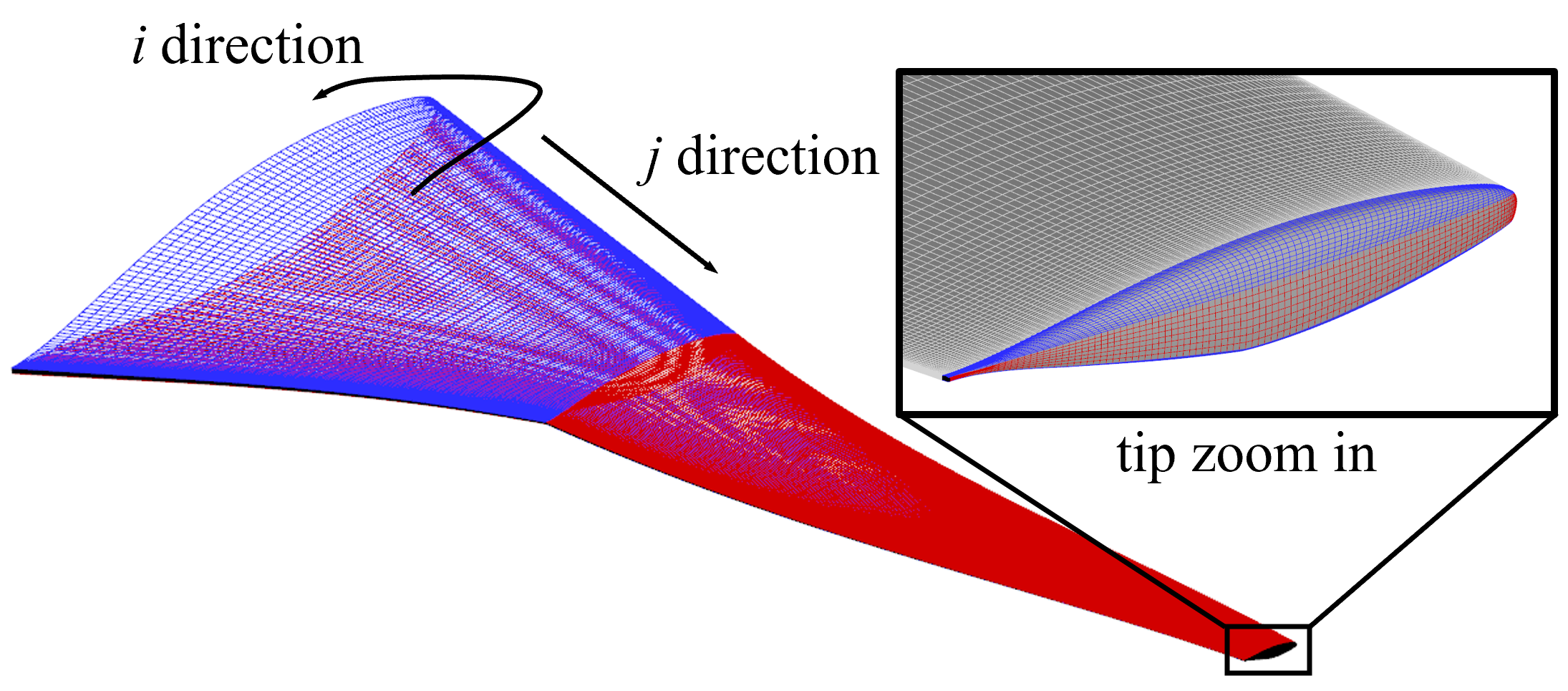}
    \caption{Surface computation mesh of the wings}
    \label{fig:wingmesh}
\end{figure}

The wing surface mesh is generated using an in-house code. As illustrated in Figure \ref{fig:wingmesh}, the wing surface is divided into two segments split at the kink, and the tips for all wing shapes are adapted from the CRM geometry \cite{lyu_crmmesh_2015} and aligned with the tip section airfoil.

Despite many validations of the \verb|ADflow| code in wing simulations, we also conducted a mesh-convergence study for the data-generation pipeline. \modified{We include the CRM under multiple transonic design points, DLR-F6, and a single-segment wing benchmark, DPW-W1} \cite{sclafani_dpw_2007}. \modified{Three mesh resolutions are generated by varying the surface mesh resolution, the volume mesh marching steps, and the surface $y+$. 

Fig.} \ref{fig:meshconverror} \modified{illustrates $C_D$ values with respect to the second-order mesh size factor $N^{-2/3}$ where $N$ is the mesh size. We present the selection of wings and operating conditions, along with results from Lyu et al.'s study} \cite{lyu_crmmesh_2015} \modified{on the CRM. The estimated M-size mesh error, $\delta_M$, was obtained by extrapolation, indicating that it remains at a similar level to that reported in the literature. The results also suggest that while different wing geometries lead to different error levels, operating conditions may have an even stronger influence. In our studied cases, higher Mach numbers tend to exhibit larger errors, likely due to stronger shock waves. Fig. }\ref{fig:meshconvshock} \modified{further shows that the sectional $C_p$  distributions with a zoom-in to the shock wave region. The shock location remains nearly unchanged, although a slightly larger but acceptable discrepancy is observed near the outboard aft-shock acceleration region.}

\begin{modifiedblock}
\begin{figure}[H]
    \small
    \centering
    \begin{subfigure}{\textwidth}
    \centering
    \begin{tikzpicture}
        \begin{axis}[
            width=0.8\textwidth,
            height=0.35\textwidth,
            xlabel={$N^{-2/3}$},
            ylabel={$C_D$ (count)},
            xmax=0.0002,
            axis lines=left,
            tick align=inside,
            axis line style={-},
            legend style={at={(0.7,0.92)},anchor=north}
        ]
            
            \addplot+[color=black, dashed, mark=o, mark size=3pt,] coordinates {(0.00000266, 199.2) (0.00001063, 199.7) (0.00004254, 201.7) (0.00017015, 211.1)};
            \node[color=black, anchor=east, xshift=-5pt, yshift=-2pt] 
                  at (axis cs:0.00017015, 211.1) {Lyu et al. \cite{lyu_crmmesh_2015}, $\delta_M=2.82$};
            
            \addplot+[color=black, mark=o, mark size=3pt,] coordinates {(0.00002324, 200.07603) (0.00004254, 201.00529) (0.00008355, 203.48084)};
            \node[color=black, anchor=west, xshift=-2pt,yshift=-10pt] 
                  at (axis cs:0.00008355, 203.48084) {CRM, $Ma=0.85, C_L=0.50$, $\delta_M=2.33$};
            
            \addplot+[color=black, mark=triangle, mark size=3pt,] coordinates {(0.00002324, 228.49492) (0.00004254, 229.53149) (0.00008355, 232.11922)};
            \node[color=black, anchor=west, xshift=2pt] 
                  at (axis cs:0.00008355, 232.11922) {CRM, $Ma=0.87, C_L=0.47$, $\delta_M=2.50$};
            
            \addplot+[color=black, mark=square, mark size=3pt,
              point meta=explicit symbolic,
              nodes near coords={\pgfplotspointmeta},
              nodes near coords style={xshift=-10pt,},
            ] coordinates {
                (0.00002324, 208.51703) [L]
                (0.00004254, 209.20443) [M]
                (0.00008355, 211.73178) [S]};
            \node[color=black, anchor=west, xshift=-2pt, yshift=10pt] 
                  at (axis cs:0.00008355, 211.73178) {CRM, $Ma=0.82, C_L=0.54$, $\delta_M=2.10$};
            
            \addplot+[color=red, mark=x, mark size=3pt,] coordinates {(0.00002324, 234.595) (0.00004254, 235.2895) (0.00008355, 237.11507)};
            \node[color=red, anchor=west, xshift=-12pt, yshift=-8pt] 
                  at (axis cs:0.00008355, 237.11507) {DPW-W1, $Ma=0.76, C_L=0.50$, $\delta_M=1.72$};
            
            \addplot+[color=blue, solid, mark=+, mark size=3pt,] coordinates {(0.00002324, 204.68842) (0.00004254, 205.39498) (0.00008355, 207.13369)}; 
            \node[color=blue, anchor=west, xshift=2pt, yshift=-4pt] 
                  at (axis cs:0.00008355, 207.13369) {DLR-F6, $Ma=0.75, C_L=0.50$, $\delta_M=1.69$};
        \end{axis}
    \end{tikzpicture}
    \caption{Errors of drag coefficients (in count), $\delta_M$ denotes the estimated error of the medium mesh}\label{fig:meshconverror}
    \end{subfigure}
    \\[5pt]
    \begin{subfigure}{\textwidth}
            \centering
            \includegraphics[width=0.85\linewidth]{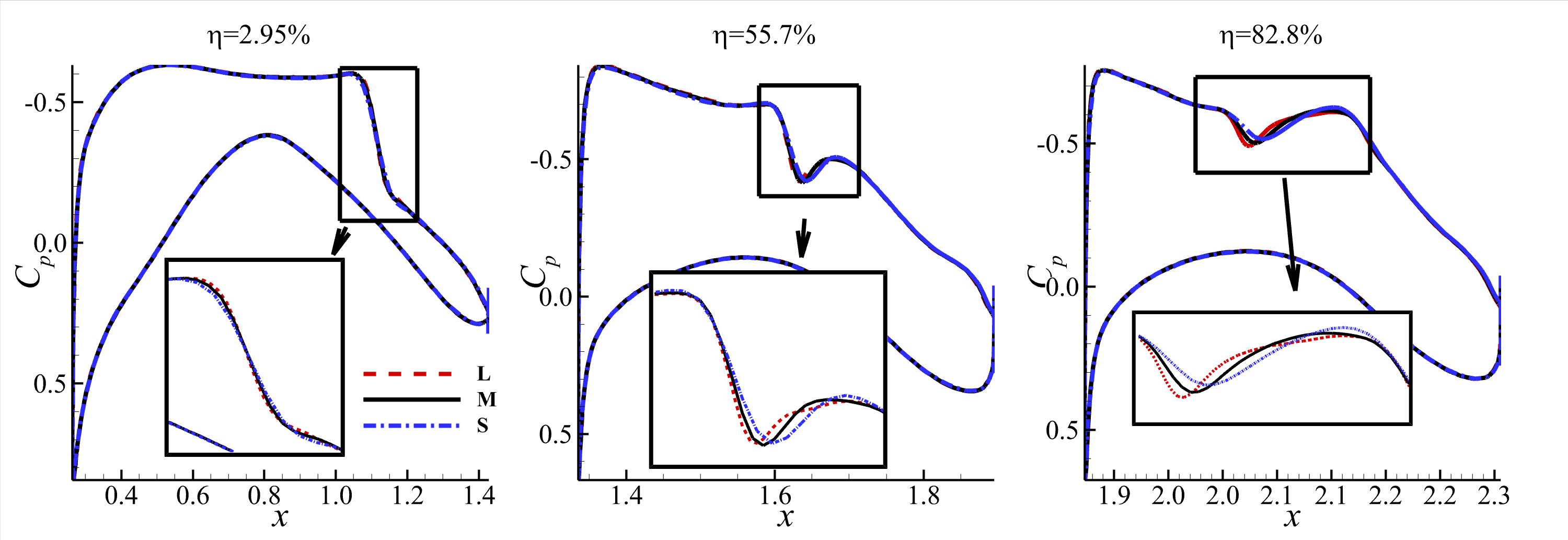}
            \caption{Pressure distribution on typical sections of CRM ($Ma=0.85, C_L=0.50$)}\label{fig:meshconvshock}
    \end{subfigure}
    \caption{Mesh convergence study of the typical wings}
    \label{fig:meshconv}
\end{figure}
\end{modifiedblock}

Considering the time and memory cost, the M-size mesh is used to generate the dataset. It has 249 cells along the airfoil circumferential direction, including \modified{9 cells across the trailing-edge height.} In the spanwise direction, it has 41 and 125 cells for the inner and outer segments. The surface mesh is refined near the leading and trailing edges in the circumferential direction and near the wing tip in the spanwise direction. The volume O-shape mesh is then extruded from the surface mesh using \verb|pyHyp| \cite{Secco2021}. It marches 81 layers with the starting layer height to fulfill $y^+\approx 1$ and ends at 50 times the root chord.

The simulations are conducted on the 160-core high-performance computing cluster at Tsinghua University for over four months. During the simulation, the lift coefficient is monitored to ensure steady-state convergence. Only cases with lift coefficient fluctuations below 0.0005 over the last 10 outer iterations are included in the final dataset. As a result, 28,856 valid wing flow fields are obtained. 

\section{Data Record}\label{sec:pp}

Table \ref{tab:files} summarizes the data that are available for direct download via Hugging Face \cite{yunjia_yang_2026}

\noindent (\href{https://huggingface.co/datasets/yunplus/SuperWing}{https://huggingface.co/datasets/yunplus/SuperWing}) . \modified{\textbf{The dataset includes wing geometries, aerodynamic coefficients, surface flow fields, and three-dimensional volumetric flow fields.} To facilitate downstream model development, the raw solver outputs that were originally stored in CFD General Notation System (CGNS) format with an Advanced Data Format (ADF) backend are post-processed into a ready-to-use point cloud representation. For the surface flow, we additionally provide a two-dimensional structured mesh version, which serves as a convenient benchmark for models that require structured inputs.}

\begin{modifiedblock}    
\begin{table}[H]
    \centering
    \small
    \caption{Data description for \textit{SuperWing} dataset}
    \label{tab:files}
    \begin{threeparttable}
    \begin{tabular}{M{1.2cm}M{3cm}M{4.5cm}M{2.5cm}c}
    \hline \hline
        Type & File & Description & Shape & Size  \\ 
        \hline
        \multirow{3}{*}{Metadata} & \texttt{config.dat} & shape parameters  & $N_{\mathrm{shape}} \times 56 $ & 5.0 MB   \\
         & \texttt{index.npy} & indexing, operating conditions, and aerodynamic coefficients &  $N_{\mathrm{sample}} \times 12 $ & 2.8 MB   \\
         & \texttt{training\_samples\_index.txt} & training sample split & -- & 0.1 MB \\\hline
        \multirow{4}{*}{\makecell{Surface \\ mesh\\ \& surface \\flow}} & \texttt{data\_surf.npy.zst} & surface simulation mesh and flow quantities on mesh points (cell center) & $N_{\mathrm{sample}} \times 10 \times 44,096 $ & \makecell{26.7 GB \\ (85.3 GB) \tnote{a}}\\
        & \texttt{origingeom.npy} & reference surface mesh (grid points) & $N_{\mathrm{shape}} \times 3 \times 129 \times 257 $ & 3.3 GB  \\
         & \texttt{geom0.npy} & reference surface mesh (cell center) & $N_{\mathrm{shape}} \times 3 \times 128 \times 256 $ & 3.3 GB  \\
         & \texttt{data.npy} & surface flow quantities at reference mesh (cell center) &  $N_{\mathrm{sample}} \times 3 \times 128 \times 256 $ &  22.7 GB  \\\hline
         Volumetric flow & \texttt{data\_vol.xx.npy.zst} & coordinates and flow quantities at near-field volumetric simulation mesh (cell center) & $N_{\mathrm{sample}} \times 8 \times 3,086,720 $~~\tnote{b} & \makecell{3.2 TB\\ (5.3 TB) \tnote{a}} \\
        \hline \hline
    \end{tabular}
    \begin{tablenotes}
        \footnotesize
        \item[a] The data files are compressed with Zstandard (\texttt{zstd}), and the original file size is in parentheses.
        \item[b] Considering the large size of the volumetric flow, it was split into 45 files by every 100 wing shapes.
    \end{tablenotes}
    \end{threeparttable}
\end{table}
\end{modifiedblock}

\subsection{Meta data}

\modified{The metadata contain the parametric description of the geometry, operating conditions, and key aerodynamic outputs for each sample. In addition, they include group identifiers and a predefined train–test split used in the experiments presented in this work.}

\paragraph{configs.dat} \modified{stores the parametric definition of each wing geometry, enabling reconstruction from scratch using the procedure described in \textit{Method} Section}. It includes:

\begin{itemize}
    \item \textbf{Global planform parameter} (\textit{\underline{Columns 1--7}}) including $\Lambda_\mathrm{LE}$, $\Gamma_\mathrm{LE,tip}$, $\Gamma_\mathrm{LE,kink}$, $AR$, $TR$, $\eta_k$, and $\kappa_\mathrm{root}$. Their definition is given in Tables \ref{tab:ranges0} and \ref{tab:ranges}.
    
    \item \textbf{Spanwise variation parameter} (\textit{\underline{Columns 8--17}}) including $r_{t} (\times 3)$, $r_{\delta} (\times 4)$, and $\alpha_{\mathrm{tw}} (\times 4)$. Their definition is given in Table \ref{tab:ranges}.
    
    \item \textbf{Baseline airfoil shape} (\textit{\underline{Columns 18--38}}) including the CST coefficients for the upper surface $(\times 10)$ and lower surface$(\times 10)$.
    
    \item \textbf{Assigned operating conditions } (\textit{\underline{Columns 39--56}}) including eight sets of $Ma$ and $\alpha$. Note that not all of them necessarily appear in the final dataset, as some cases may fail to converge during CFD simulation.
    
\end{itemize}
   
\paragraph{index.npy} \modified{provides essential metadata for model training, where each row corresponds to a single flow-field sample. The included variables are:}

\begin{itemize}
    \item \textbf{Mapping between samples and wing geometries}. Since multiple operating conditions are associated with each geometry, \textit{\underline{Column 1}} specifies the geometry index (corresponding to \texttt{configs.dat}), and \textit{\underline{Column 2}} indicates the operating condition index for that geometry and  \textit{\underline{Column 2}} provides the index of operating condition counted in each wing shape.
    
    \item \textbf{Operating conditions}. \textit{\underline{Column 3}} stores the angle of attack ($\alpha$), and \textit{\underline{Column 4}} the Mach number ($Ma$).

    \item \textbf{Reference quantities} used for coefficient non-dimensionalization. \textit{\underline{Column 5}} gives the half reference area ($S_{1/2}$), defined here as the projected planform area (the region $O'DFGEB'$ in Fig. \ref{fig:three-view}), and \textit{\underline{Column 6}} provides the half span ($b_{1/2}$).

    \item \textbf{Aerodynamic coefficients}, which are critical targets for many downstream tasks. We include the lift, drag, and $z$-axis pitching moment coefficients ($C_L$, $C_D$, $C_{M,z}$). These are computed from the surface pressure coefficient ($C_p$) and skin-friction coefficient vector $C_{\bm f} = [C_{f,x}, C_{f,y}, C_{f,z}]^T$. Note that they are already non-dimensionalized by freestream dynamic pressure $0.5 \rho_\infty V_\infty^2$. 

    The force coefficients are first obtained by integrating surface quantities in the mesh coordinate system:
    
    \begin{equation}\label{eqn:force}
        C_{\bm{F}}=[C_{x}, C_{y}, C_{z}]^T=\frac{1}{S_{1/2}}\sum_{i=1}^{N_{\mathrm{cell}}}\left[C_{p}^{(i)}\bm n^{(i)}+\left(C_{\bm f}^{(i)}-\left(C_{\bm f}^{(i)}\cdot \bm n^{(i)}\right)\bm n^{(i)}\right)\right]A^{(i)} ,
        \end{equation}
        where $\bm n^{(i)}$ and $A^{(i)}$ denote the outward normal vector and area of the $i$-th surface cell, respectively.
        
        The lift and drag coefficients are then obtained by rotating the force vector according to the angle of attack:
        \begin{equation}
            \left[C_L, C_D, C_z\right]^T = \bm R_\alpha C_{\bm F}.
        \end{equation}
        
        The $z$-axis pitching moment coefficient is defined as
        \begin{equation}
        \label{eqn:moment}
        C_{M,z}=\frac{1}{S_{1/2}c_{\textrm{ref}}}\sum_{i=1}^{N_{\mathrm{cell}}}\bm r^{(i)}\times\left(\left[C_{p}^{(i)}\bm n^{(i)}+\left(C_{\bm f}^{(i)}-\left(C_{\bm f}^{(i)}\cdot \bm n^{(i)}\right)\bm n^{(i)}\right)\right]A^{(i)}\right)
    \end{equation}
    where $\bm r^{(i)}$ is the position vector from the reference point to the center of the $i$-th surface cell, $c_{\textrm{ref}} = 1.0$ is the reference chord, and $\bm e_z$ is the unit vector along the $z$ axis. The origin is used as the moment reference point, and a nose-down pitching moment is defined as positive according to our coordinate system.

    In \texttt{index.npy}, we provide two sets of aerodynamic coefficients. The first set (\textit{\underline{Columns 7--9}}) is computed directly on the original CFD surface mesh using the \texttt{ADflow} solver. The second set (\textit{\underline{Columns 10--12}}) is evaluated on a structured reference mesh, primarily for machine learning applications.
    
\end{itemize}

\paragraph{training\_samples\_index.txt} \modified{stores the indices of the \textit{training samples} in our following technical validation. To fairly assess the model's generalization performance across wing shapes, we split the dataset by shape. Specifically, we randomly select 90\% of the wing shapes, and the corresponding samples are recorded in \texttt{training\_samples\_index.txt} and are used for training. The remaining samples are for testing and remain unseen during model training.}

\paragraph{parquet files} \modified{(\texttt{train.parquet}, \texttt{test.parquet}) provide metadata organized by training and test data, enabling easy visualization in HuggingFace. }

\subsection{Surface shape and surface flow}

\modified{Instead of using from-scratch shape parameters, we recommend describing wing shapes using a discretized wing surface. It is agnostic to the parameterization method, meaning a model trained with it can be transferred to wing shapes with other parameterizations or more complex configurations. The SuperWing dataset provides two types of discretized wing surface: the point cloud one is exactly the simulation mesh, and the other is an interpolated reference mesh for the structured-input model. The surface flow is also provided on the surface mesh points.}

\paragraph{Original data} \modified{(\texttt{data\_surf/data\_surf.npy.zst}) provides the centric coordinates of the exact surface mesh for the simulations, and the surface flow values on the surface mesh centers. Nine channels described in Table} \ref{tab:channels} \modified{are provided at each point. Besides the coordinates, we have}

\begin{itemize}
    \item Pressure coefficient
        \begin{equation}
            C_p = \frac{p-p_\infty}{0.5\rho_\infty V_\infty^2}
        \end{equation}
    \item Skin-friction coefficient vector 
        \begin{equation}
            C_{\bm f} = [C_{f,x}, C_{f,y}, C_{f,z}]^T = \frac{\bm{\tau_w}}{0.5\rho_\infty V_\infty^2}, \quad \bm{\tau_w}=\mu \frac{\partial \bm {u_t}}{\partial s_n}
        \end{equation}
        where $\bm{u_t}$ is the tangential velocity at the wall surface, while $s_n$ is the wall-normal coordinate.
    \item Density, and wall temperature relative to the freestream values
        \begin{equation*}
            \tilde \rho = \frac{\rho}{\rho_\infty}, \tilde T = \frac{T_w}{T_\infty}
        \end{equation*}
\end{itemize}

\modified{To enable compact storage, we flatten the original multi-block structured (Fig.} \ref{fig:surfaceframe}) \modified{mesh into a one-dimensional sequence of 44,096 points. A visualization of the point cloud is displayed in Fig. }\ref{fig:surfacepointcloud} \modified{with 8\% of the points and colored with the pressure coefficient $C_p$.}

\begin{modifiedblock}
\begin{figure}[H]
    \centering
    \begin{subfigure}{0.4\textwidth}
        \centering
        \includegraphics[width=\linewidth]{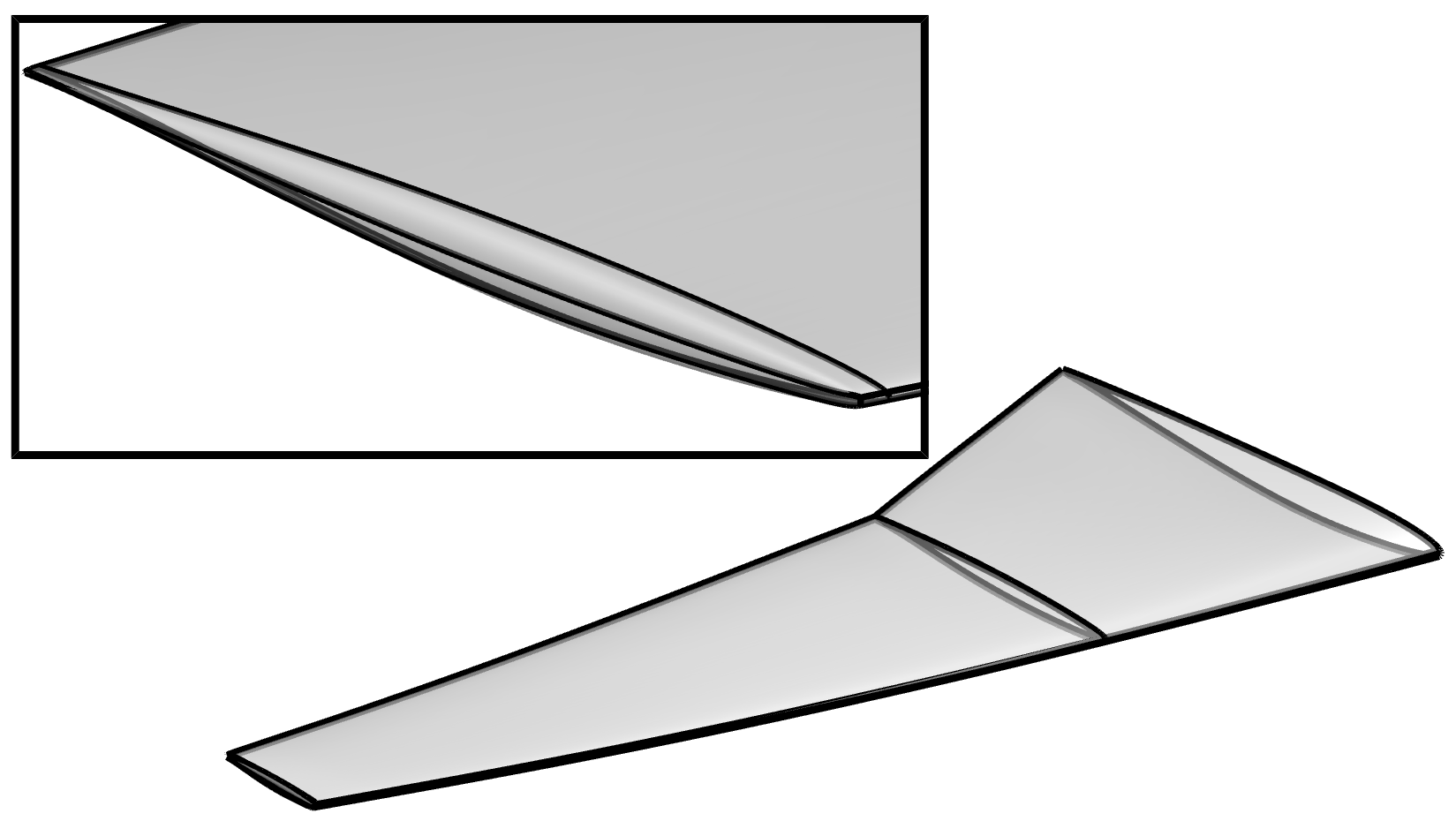}
        \caption{Original surface mesh blocks}
        \label{fig:surfaceframe}
    \end{subfigure}
    \begin{subfigure}{0.5\textwidth}
        \centering
        \includegraphics[width=\linewidth]{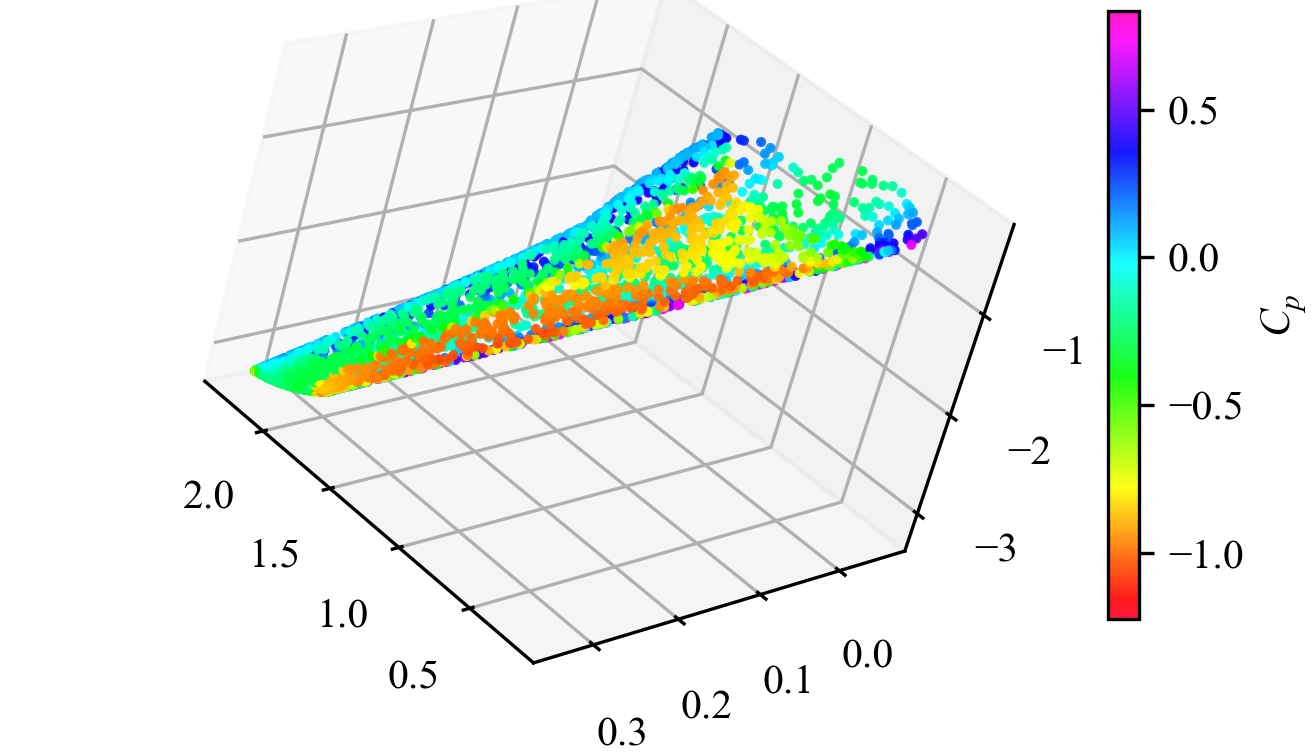}
        \caption{Visualization of surface point cloud}
        \label{fig:surfacepointcloud}
    \end{subfigure}
    \caption{Visualization of wing shapes and surface point cloud}
    \label{fig:surface}
\end{figure}
\end{modifiedblock}

\paragraph{Structured surface shape and flow} \label{sec:refmesh}(\texttt{origingeom.npy}, \texttt{geom0.npy}, \texttt{data.npy}) ~~Besides the raw multi-block solver output, we also prepare the surface mesh and flow fields in a format suitable for ML models with structured inputs and outputs.

The simulation mesh on the wing surface is first interpolated to a \textit{reference mesh}. In the spanwise, \modified{the end surface at the wing tip, together with a very small neighboring region, is removed during post-processing. This allows the reference mesh to be unfolded, as shown in Fig.} \ref{fig:unfold} \modified{, thereby enabling the construction of a consistent structured mesh. The removed region accounts for approximately $\approx 0.3\%$ of the span. When reconstructing the aerodynamic coefficients, the contribution of this region is recovered by interpolating the flow quantities to the corresponding streamwise locations on the upper and lower surfaces. For the remaining part, the reference mesh locates 129 evenly spaced points in the spanwise direction.} For each location, a fixed set of normalized chordwise positions $\{(x/c)_i\}$ is used for both the upper and lower surfaces, and the tail edge height is represented only with one cell. The procedure produces a final reference grid of $257 \times 129$ points per wing, stored in \texttt{origingeom.npy}. Since flow variables are stored at cell centers, we also provide a cell-centered reference mesh, \texttt{geom0.npy}. The flow quantities are interpolated to this mesh and stored in \texttt{data.npy}. Validation shows that excluding the tip region and interpolating onto the reference grid introduces an error of less than 0.1\% in the global aerodynamic coefficients. \modified{We also provide a comparison of the surface pressure coefficient before and after interpolation in Fig.} \ref{fig:interperror}, \modified{demonstrating that the shock wave is accurately preserved.

Given their importance, the surface flow data here include only pressure and friction coefficient. The latter is further decomposed into a streamwise part $C_{f,\tau}$ in the $x$-$y$ plane, and a spanwise part $C_{f,z}$, in the $z$-direction. For $C_{f,\tau}$, the positive direction is defined from the lower to the higher $i$ index, which means if there is no inverse flow, $C_{f,\tau}$ is positive on the upper surface and negative on the lower surface. Because these three variables have different magnitudes, we obtain a representative channel range, $r_X,~ X\in \left[C_p, C_{f,\tau}, C_{f,z}\right]$, by first computing the sample-wise range for each variable and averaging them across the dataset. }

\begin{modifiedblock}
\begin{equation}\label{eqn:rx}
    r_X = \frac{1}{N_s}\sum_1^{N_s}\left(\max_{i,j}X^\mathrm{CFD}_n-\min_{i,j}X^\mathrm{CFD}_n\right), \quad X\in \left[C_p, C_{f,\tau}, C_{f,z}\right]
\end{equation}
\end{modifiedblock}

\modified{Based on these ranges, we apply scaling factors of 1, 150, and 300, for $C_p$, $C_{f,\tau}$, and $C_{f,z}$, respectively, so that the three channels have comparable magnitudes.}

\begin{modifiedblock}
\begin{figure}[H]
    \centering
    \begin{subfigure}{0.55\textwidth}
        \centering
        \includegraphics[width=\linewidth]{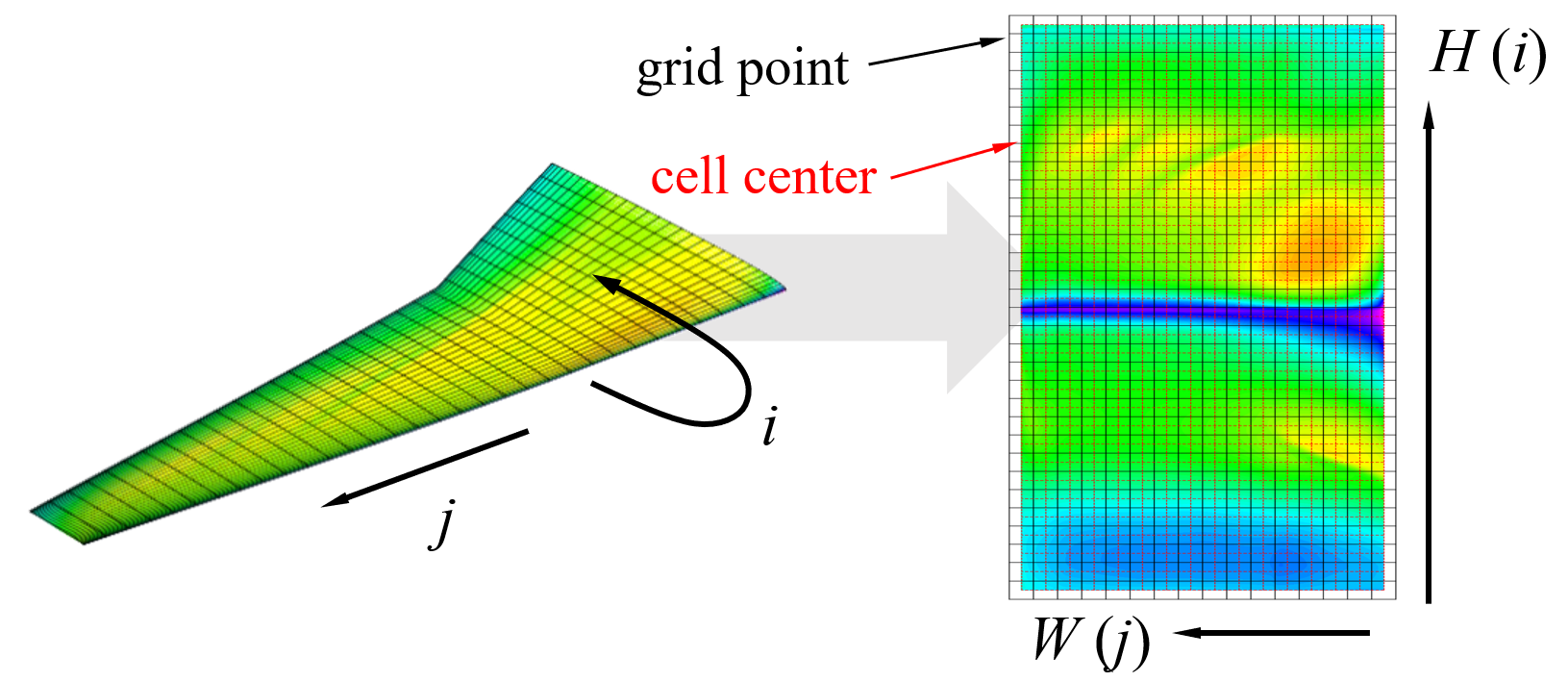}
        \caption{Unfolding procedure}\label{fig:unfold}
    \end{subfigure}
    \begin{subfigure}{0.4\textwidth}
        \centering
        \includegraphics[width=\linewidth]{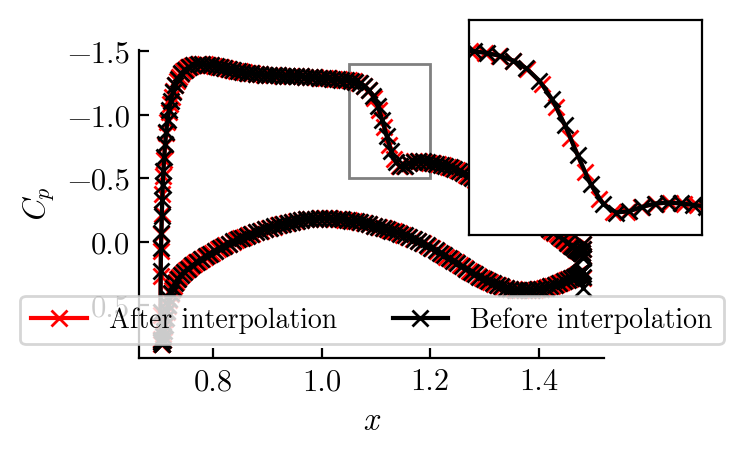}
        \caption{Comparison of $C_p$ before and after interpolation}\label{fig:interperror}
    \end{subfigure}
    \caption{Transferring surface mesh and quantities from simulation mesh to reference mesh}
\end{figure}
\end{modifiedblock}

\subsection{Volumetric flow}

Although surface flow data are sufficient to predict the surface flow and aerodynamic coefficients crucial to design optimization, volumetric flow data are also provided for potential future use. 

\paragraph{data\_vol/data\_vol.xx.npy.zst} \modified{provides the volumetric flow, including the cell-centric coordinates and five core flow quantities: density, pressure, and the three velocity components at each simulation cell. They are again defined as the relative value to the freestream, and their order is shown in Table } \ref{tab:channels}. \modified{Similar to the surface data, we flatten and concatenate the multi-block mesh, as shown in Fig.} \ref{fig:volmesh}, \modified{into a one-dimensional point cloud sequence. Since the volumetric mesh is generated by marching outward from the wing surface to the far field, each volumetric block corresponds to one surface block and has an additional dimension for wall-normal marching direction. Given that it requires a large far field to implement the freestream boundary condition in simulation, the flow variables at far-field mesh points show only negligible deviations from the freestream values. To avoid this redundancy, we include the first 71 layers of mesh in the wall-normal dimension for each block. This produces 3,086,720 points per volumetric flow field.}

\begin{modifiedblock}
\begin{figure}[H]
    \centering
    \includegraphics[width=0.5\linewidth]{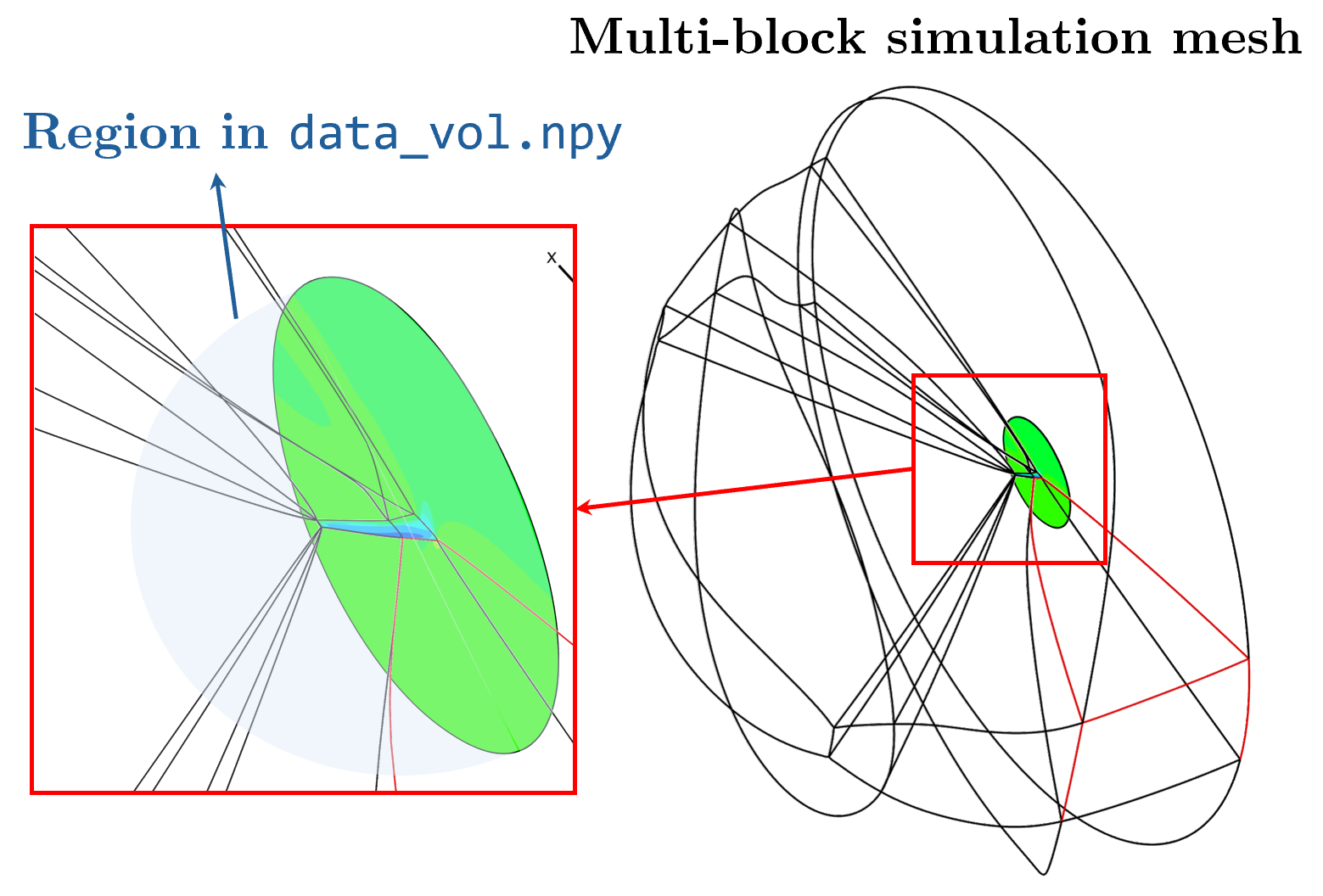}
    \caption{Visualization of volumetric data}
    \label{fig:volmesh}
\end{figure}

\begin{threeparttable}[H]
    \small
    \centering
    \caption{Physical channels description of the surface and volumetric flow}
    \label{tab:channels}
    \begin{tabular}{ccc|cc}
    \hline \hline
        & \texttt{data\_surf.npy} & \texttt{data\_vol.npy} &  & \makecell{\texttt{origingeom.npy}, \texttt{geom0.npy}}\\
        \midrule
        0 & \multicolumn{2}{c|}{Coordinate $x$} & 0 & Coordinate $x$\\
        1 & \multicolumn{2}{c|}{Coordinate $y$} & 1 & Coordinate $y$\\
        2 & \multicolumn{2}{c|}{Coordinate $z$} & 2 & Coordinate $z$\\ 
        3 & \multicolumn{2}{c|}{Density $\tilde \rho$} & & \texttt{data.npy}\\
        4 & Pressure coef. $C_{p}$ & Pressure $\tilde p$ & 0 & Pressure coef. $C_{p,\mathrm{scaled}}$\\
        5 & $x$ skin friction coef. $C_{f,x}$ & $x$ velocity $\tilde V_x$ & 1 & Streamwise skin friction coef. $C_{f,\tau,\mathrm{scaled}}$\\
        6 & $y$ skin friction coef. $C_{f,y}$ & $y$ velocity  $\tilde V_y$ & 2 & $z$ skin friction coef. $C_{f,z,\mathrm{scaled}}$\\
        7 & $z$ skin friction coef. $C_{f,z}$ & $z$ velocity  $\tilde V_z$\\
        8 & Temperature $\tilde T$ & \\
        \hline \hline
    \end{tabular}
    \begin{tablenotes}
        \item[a] $\tilde{(\cdot)}$ means the relative value to the freestream condition ($\rho_\infty, p_\infty, T_\infty, V_\infty$).
    \end{tablenotes}
\end{threeparttable}
\end{modifiedblock}

\section{Data Overview}

\begin{figure}[H]
    \centering
    \includegraphics[width=1\linewidth]{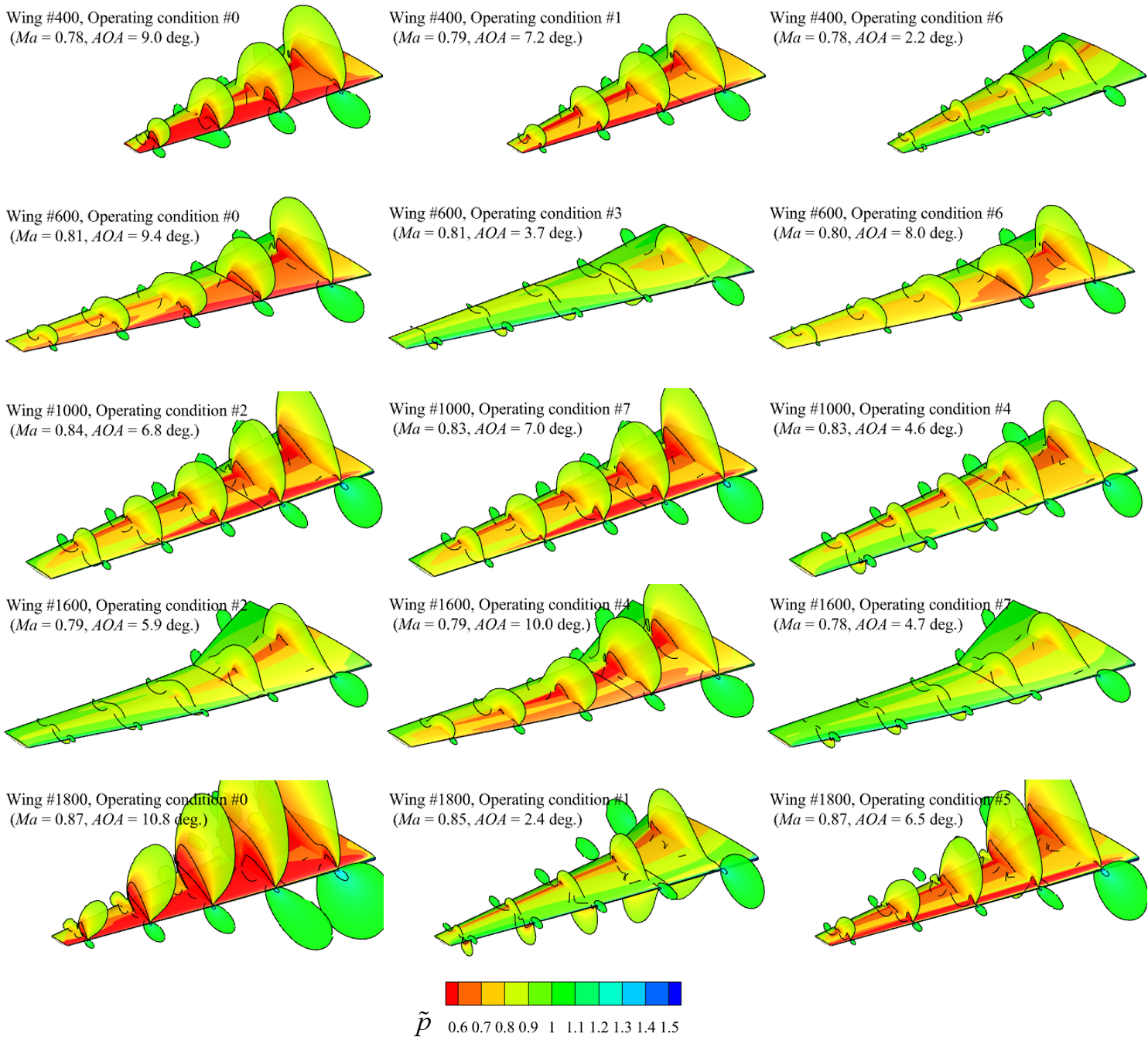}
    \caption{Visualization of flow fields around wings in the dataset}
    \label{fig:wingff}
\end{figure}

Fig. \ref{fig:wingff} demonstrates the $C_p$ fields on the surface of and around several wings in the dataset, where complex flow structures such as multiple shock waves can be observed.

\section{Technical Validation}

This section focuses on validating the sampling method, demonstrating that a machine-learning model trained on the provided dataset has the potential to be applied to downstream design optimization tasks. Specifically, we are predicting wing surface flow from wing shapes and operating conditions. The predictive accuracy and computational cost performance of two Transformer-based models are first compared with that of the commonly used U-Net \cite{navab_u-net_2015}. Later, the pre-trained model is utilized to predict two typical wings, DLR-F6 and CRM, which are outside the training distribution. 

\subsection{Experimental setup}

\subsubsection{Problem definition}
Three models are set up with the same prediction task for the surface flow of transonic wings. The input is the reference mesh, $\bm g$ (\verb|geom0.npy|), which has a size of $256 \times 128 \times 3$. \modified{The operation conditions $\bm c = [\alpha, Ma]$ are expanded to the same spatial shape as the reference mesh, and concatenated with it before being entered into the model.}

The primary output is the surface flow on the same reference mesh (\verb|data.npy|) that includes the surface pressure and friction coefficient distributions. 

\subsubsection{Model details}\label{sec:baseline}

\paragraph{U-Net} 
We employ the U-Net architecture from our previous work on wing flow field prediction \cite{yang_rapid_2025} as the baseline model. It consists of a symmetric encoder-decoder structure as shown in Fig. \ref{fig:unet}, and operating conditions are concatenated with the latent representation between the encoder and decoder layers. Both the encoder and decoder consist of six ResNet layers. Each ResNet layer includes a residual block for down- or up-sampling that halves or doubles the $i$-direction resolution, followed by a standard residual block. The number of hidden dimensions are $(16, 32, 32, 64, 64, 128)$ for encoder and reversed for decoder. The decoder concludes with a final convolutional layer that reduces the hidden channels. 

\paragraph{Vision Transformer (ViT)} 

The original ViT framework proposed in \cite{dosovitskiy_vit_2021} is used as shown in Fig. \ref{fig:vit}. It slices the input mesh with patching size $p = 4$, which leads to 2048 tokens in the sequence. The operating conditions are introduced by expanding them to the mesh size and input to the embedding layer together. A learnable matrix is used as the positional embedding, which is added element-wise to the input token. The Transformer backbone of ViT has 5 layers, each with 8 attention heads, 256 hidden dimensions of tokens, and an MLP ratio of 4. The output surface flow is obtained by combining the Transformer output sequence.

\begin{modifiedblock}
\begin{figure}[H]
    \centering
    \begin{subfigure}{0.3\textwidth}
        \centering
        \includegraphics[width=\linewidth]{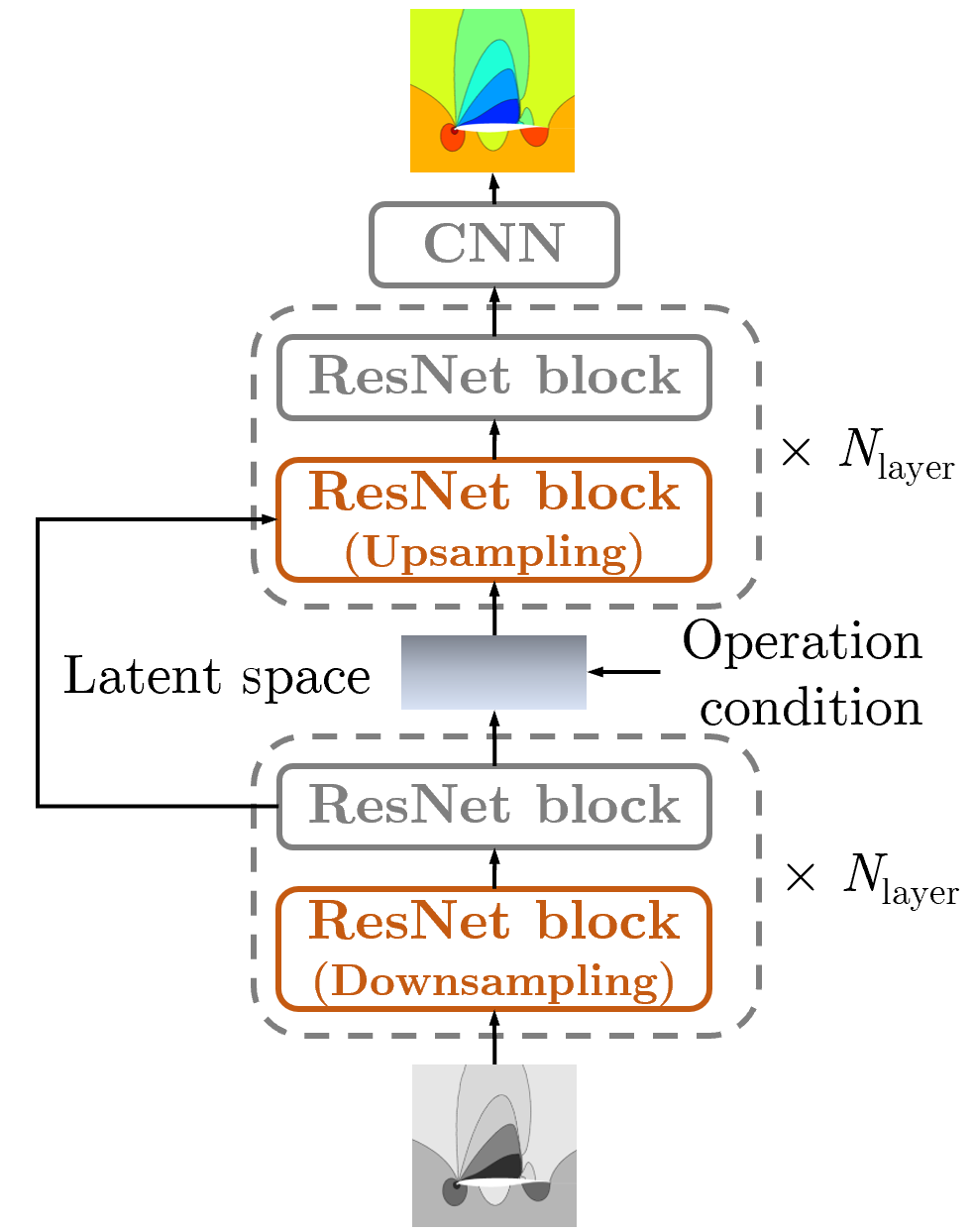}
        \caption{U-Net}\label{fig:unet}
    \end{subfigure}
    \hfill
    \begin{subfigure}{0.6\textwidth}
        \centering
        \includegraphics[width=\linewidth]{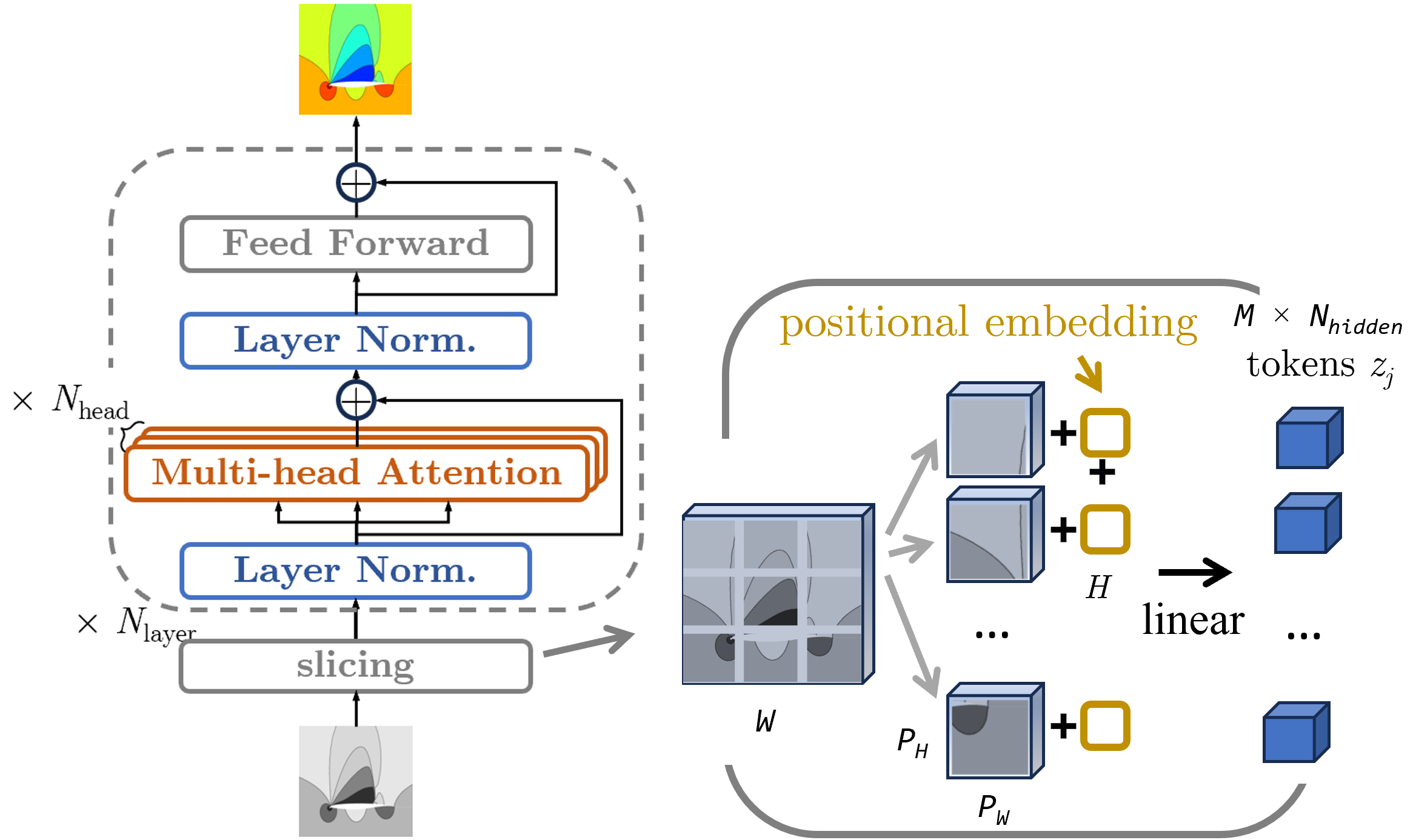}
        \caption{ViT}\label{fig:vit}
    \end{subfigure}\\[10pt]
    \begin{subfigure}{1.0\textwidth}
        \centering
        \includegraphics[width=\linewidth]{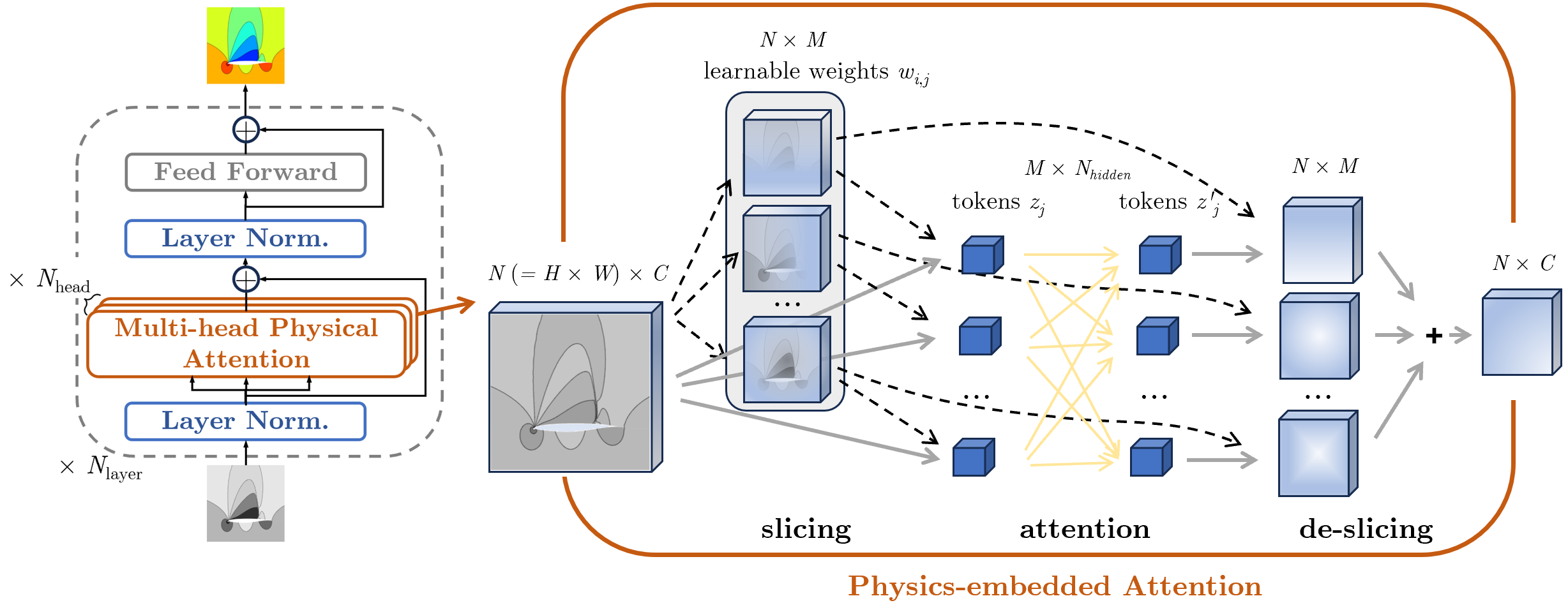}
        \caption{Transolver}\label{fig:trans}
    \end{subfigure}
    \caption{Schematic diagrams of models}
\end{figure}
\end{modifiedblock}

\paragraph{Transolver} 

Transolver \cite{wu_transolver_2024} uses a physics-embedded strategy for attention calculation \modified{, as shown in Fig.} \ref{fig:trans}. \modified{Transolver operates on the full-resolution point-wise representation throughout the network. Before each physics-attention stage, the full set of points in physical space is mapped to a set of latent physics tokens using learnable mapping weights derived from the input. Attention is then computed among these latent physics tokens, after which the latent information is mapped back to the original point-wise features. In this work, the Transolver's setting follows} \cite{wu_transolver_2024}. It also has 5 layers, 8 attention heads, 256 token dimensions, and an MLP ratio of 4. The number of tokens is 32. The operating conditions are processed in a manner similar to that of ViT.

\subsubsection{Training setup}

The training loss function is the mean squared error between model predictions and the ground-truth surface flow. The Adam optimizer and a mini-batch size 4 are selected based on the memory limit. A learning rate schedule based on the one-cycle policy is used to update parameters. Specifically, the learning rate increases from $4 \times 10^{-5}$ to $1 \times 10^{-3}$ in the first half of training and then decreases back to $1 \times 10^{-6}$ in the second half. The gradient clipping based on the exponential moving average (EMA) of gradients is adapted to avoid the spikes in the learning curve.

\subsection{Model performance on the \textit{SuperWing} dataset}

To ensure robustness, the training process is repeated three times, using different random splits of 90\% of the training data to cross-validate the results, and the error scores in Table \ref{tab:modperf} are the averages across the three cross-validation runs. \modified{For each sample, the predicted outputs are first converted back to the physical values by removing the scaling factors. Then, a relative error for each channel is defined with the channel range $r_X$ in Equation }\ref{eqn:rx} as:

\begin{equation}
    \delta X = \frac{1}{N_s} \sum_{n=1}^{N_s}\frac{\frac{1}{H\cdot W}\sum_{i,j}\left|X_n^\mathrm{model}-X_n^\mathrm{CFD}\right|}{r_X}, \quad X\in \left[C_p, C_{f,\tau}, C_{f,z}\right]
\end{equation}

In addition, MAEs in the lift and drag are also evaluated as: 

\begin{equation}
    \delta X = \frac{1}{N_s} \sum_{n=0}^{N_s}\left|X^\mathrm{model}-X^\mathrm{CFD}\right|, \quad X\in \left[C_L, C_D\right]
\end{equation}

\modified{The experiments are conducted on one NVIDIA RTX 4090 GPU. We also include the training time and peak memory in the table to demonstrate the scalability.}

\begin{table}[htbp]
\small
    \centering
    \caption{\label{tab:modperf}Model performance comparison}
    \begin{tabular}{ccccccccc}
        \hline\hline
        \multirow{2}{*}{Model} & \multicolumn{5}{c}{Errors} & \multirow{2}{*}{Parameter} & Time\ & Memory \\ \cmidrule{2-6}
        & \makecell{$C_p$\\$(\%)$} & \makecell{$C_{f,\tau}$\\$(\%)$} & \makecell{$C_{f,z}$\\$(\%)$} & \makecell{$C_L$\\$(\times10^{-3})$} & \makecell{$C_D$\\$(\times10^{-4})$}  & & (hours) & (GB)\\
        \midrule
        UNet & 1.101 & 0.642 & 0.698 &  23.41 & 14.78 &  9.2M & 17.1 & 5.55\\
        ViT & \textbf{0.329} & \textbf{0.245} & \textbf{0.281} &  2.76 & \textbf{2.48}  & 4.5M & 12.6 & 5.78 \\
        Transolver & 0.359 & 0.271 & 0.310 &  \textbf{2.71} & 2.53  & 3.8M & 37.9 & 16.42 \\\hline\hline
    \end{tabular}
\end{table}

With the attention mechanism, Transformer-based models significantly improve prediction accuracy despite having fewer trainable parameters. The training time is also reduced for the ViT model, with a little increase in running memory. \modified{Between the two Transformer-based models, Transolver more than doubled the training time and memory cost without much improvement in accuracy. Compared to patch-based ViT, Transolver maintains full resolution throughout the path. This leads to a larger memory footprint, thereby limiting the number of tokens it can achieve. Considering Transolver was initially designed for unstructured shape representation, ViT is recommended as the best backbone for further development of this task.}

We use three wings in the testing dataset here to visually demonstrate ViT's prediction. Fig. \ref{fig:testwing} shows the surface pressure $C_p$ and friction magnitude $||\bm {C_f}||$ contours, while Fig. \ref{fig:testwingeta} shows $C_p$ and $C_{f,\tau}$ profiles of cross sections at spanwise stations $\eta =$ 0.2, 0.5, and 0.8. \modified{Note that $C_{f,\tau}$ on the lower surface is by definition with a negative value when no inverse flow occurs. The first wing is the worst sample, with the largest errors in both $C_p$ and $C_{\bm{f}}$, while the other two are typical samples with different shock wave patterns. The results show that even in the worst case, the dominating flow structures are well captured.}

\begin{modifiedblock}
\begin{figure}[H]
    \centering
    \begin{subfigure}{1\textwidth}
        \centering
        \includegraphics[width=0.495\linewidth]{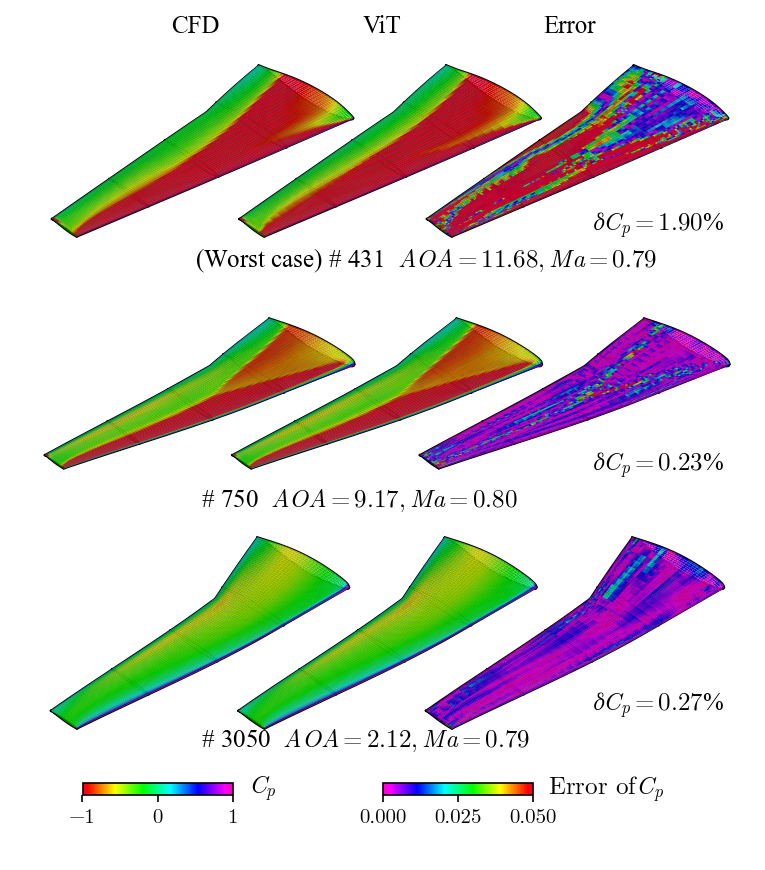}
        \hfill
        \includegraphics[width=0.495\linewidth]{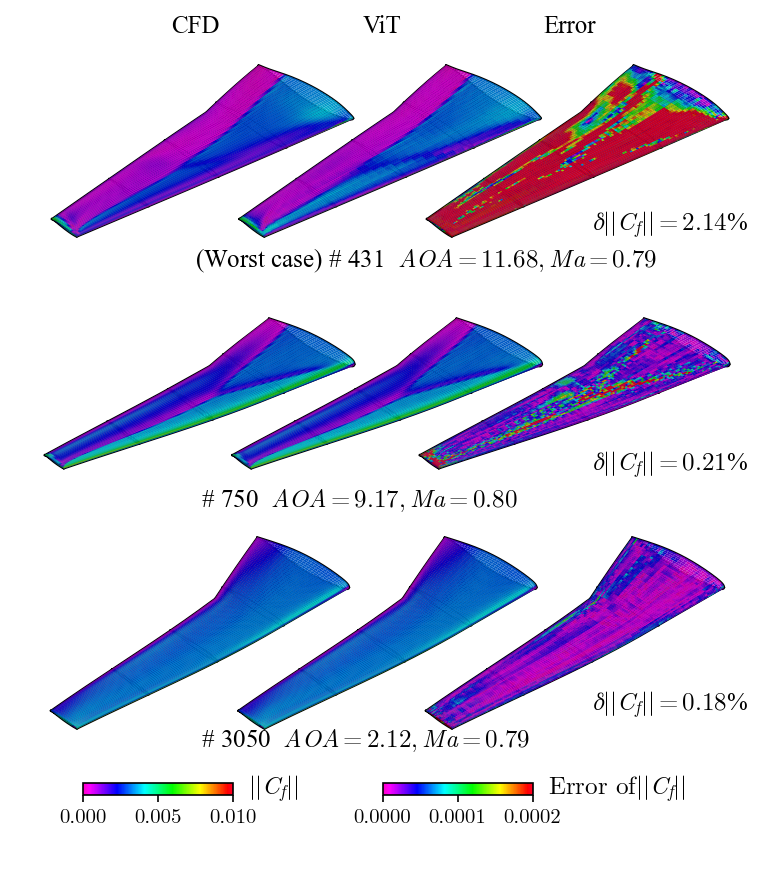}
        \caption{Surface contours}
        \label{fig:testwing}
    \end{subfigure}

    \begin{subfigure}{1\textwidth}
        \centering
        \includegraphics[width=0.40\linewidth]{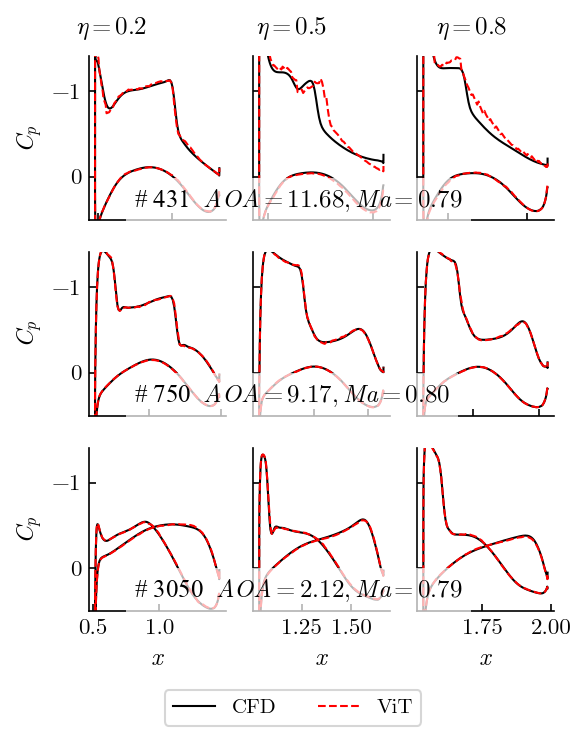}
        \includegraphics[width=0.42\linewidth]{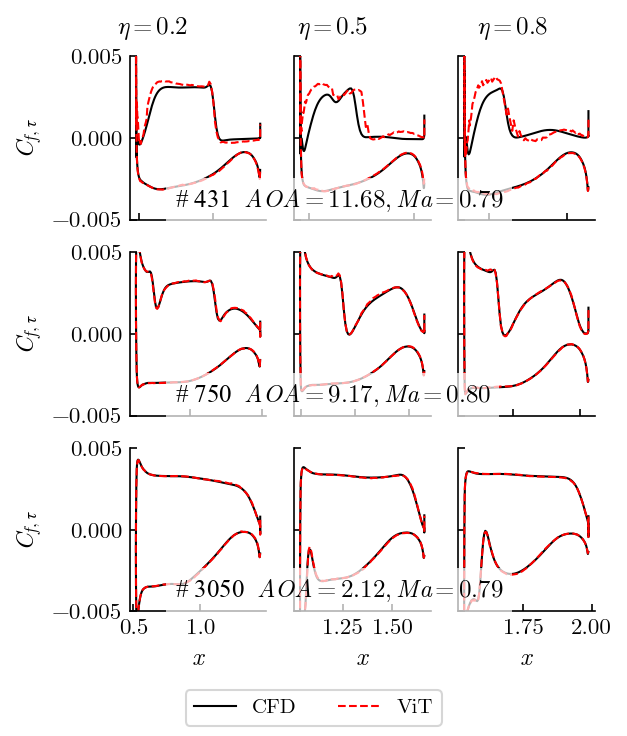}
        \caption{Cross-sectional coefficients distribution}
        \label{fig:testwingeta}
    \end{subfigure}
        
    \caption{Surface flow prediction of ViT for three wings in the test dataset }
    \label{fig:test}
\end{figure}
\end{modifiedblock}

\subsection{Model performance on out-of-distribution wing shapes}

\modified{The two more complex wing shapes, DLR-F6 and CRM, are used to demonstrate the representativeness and diversity of \textit{SuperWing} dataset. They both have several different sectional airfoil shapes, whereas the dataset generates wings from a single set of baseline CST coefficients. Specifically, we use seven sectional profiles to describe the CRM geometry and three for the DLR-F6 geometry. The profiles are directly obtained from the original geometry, and the wing simulation mesh is constructed by connecting them. After simulations, the solutions are mapped to the same structured reference mesh as the \textit{SuperWing} samples for model testing. Since the sectional profiles of these benchmark wings vary much more strongly from root to tip than those in the \textit{SuperWing} distribution, these cases provide a challenging out-of-distribution test for evaluating the generalization capability of models trained on it.}

Figures \ref{fig:testbenchf6} and \ref{fig:testbenchcrm} show the surface flow and aerodynamic coefficients predicted directly by the ViT trained with the \textit{SuperWing} dataset. \modified{On the upper surface, larger discrepancies are observed in the inner-wing region and at lower angles of attack. The predicted shock locations remain generally good, while the shock strength shows larger errors. This behavior is mainly attributed to the stronger spanwise variation of sectional profiles at the inner segment of both benchmark wings. Moreover, at lower angles of attack, the shock wave can exhibit a lambda-shock pattern, leading to a more complex flow. Nevertheless, the overall trends and integrated aerodynamic coefficients are still well captured, indicating that the model retains reasonable predictive capability on these challenging benchmark geometries.}

\section*{Data Availability}

All data are available at \href{https://huggingface.co/datasets/yunplus/SuperWing}{https://huggingface.co/datasets/yunplus/SuperWing}.

\section*{Code Availability}

The code package from \verb|MDOLab| is used for volume mesh and CFD simulation. Specifically, the code is run with the Docker image \verb|mdolab:u20-gcc-ompi-stable| that is available on \href{https://hub.docker.com/}{Docker Hub}. For data postprocessing and test model training, we provide our codes on GitHub repositories \\\href{https://github.com/YangYunjia/cfdpost}{https://github.com/YangYunjia/cfdpost} and \href{https://github.com/YangYunjia/floGen}{https://github.com/YangYunjia/floGen}, respectively.

\section*{Acknowledgments}

The authors would like to thank Zizhou He and Runze Li from Tsinghua University for their helpful advice. During writing, AI is used to polish some parts of this work. 

\section*{Author contributions}

\textbf{Yunjia Yang}: Conceptualization (equal); Data curation (lead); Methodology (lead); Visualization (lead); Writing—original draft (lead). \textbf{Weishao Tang}: Conceptualization (equal); Data curation (supporting). \textbf{Mengxin Liu}: Data curation 
(supporting).\textbf{Nils Thuerey}: Writing—review \& editing (equal). \textbf{Yufei Zhang}: Funding acquisition (supporting); Writing—review \& editing (equal). \textbf{Haixin Chen}: Funding acquisition (lead); Supervision (equal); Writing—review \& editing (equal).

\section*{Competing interests}

The authors declare no competing interests.

\section*{Funding}

This work was supported by the National Natural Science Foundation of China (NSFC) Nos. 12202243, 12372288, 12388101, and U23A2069.

\begin{modifiedblock}
\begin{figure}[H]
    \centering
    \begin{subfigure}{0.495\textwidth}
        \centering
        \includegraphics[width=\linewidth]{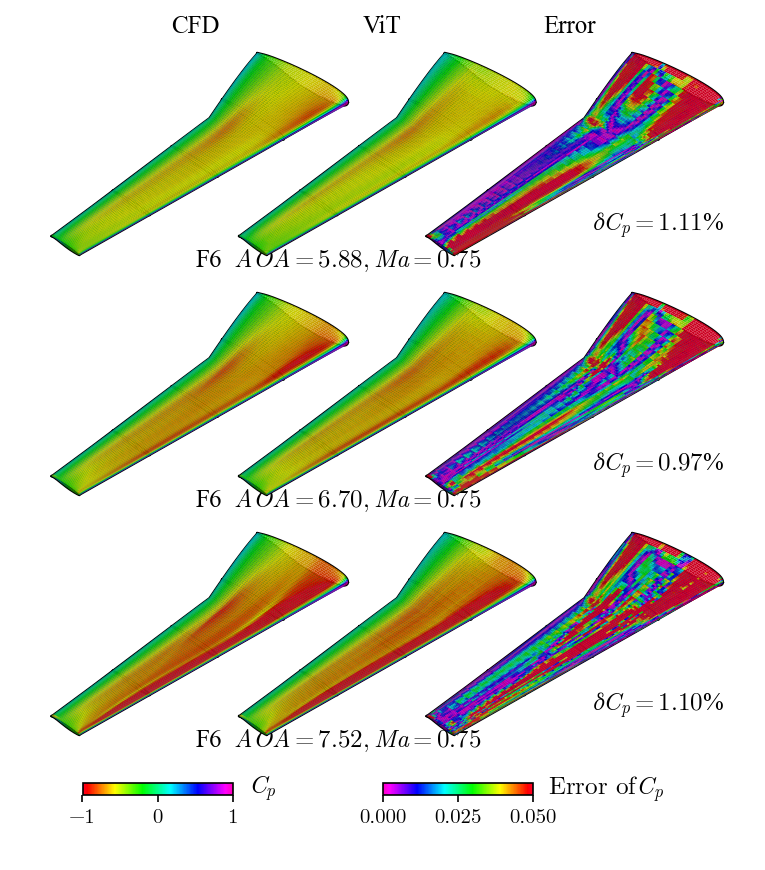}
        \caption{Surface contours}
    \end{subfigure}
    \begin{subfigure}{0.4\textwidth}
        \centering
        \includegraphics[width=\linewidth]{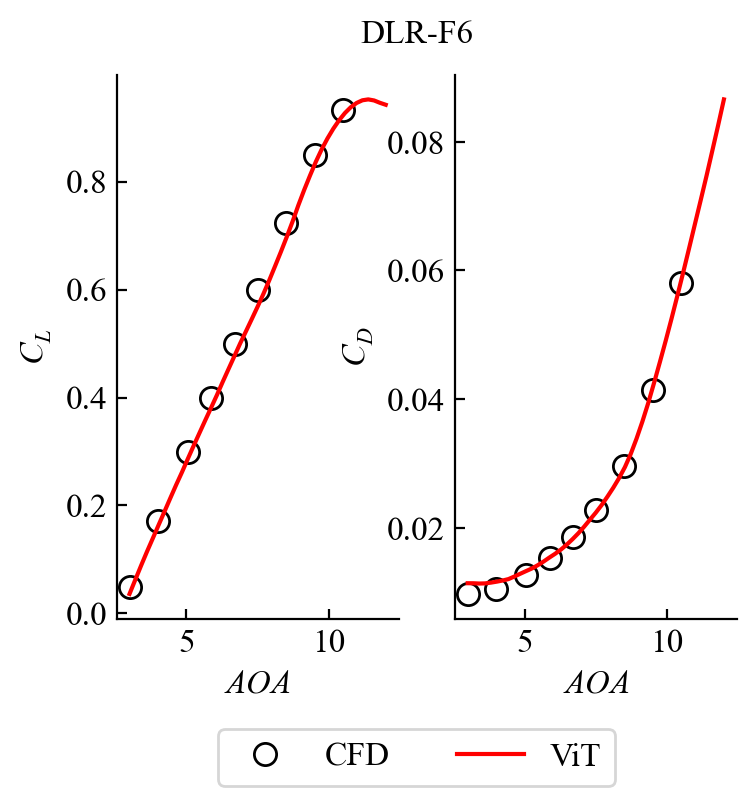}
        \caption{Aerodynamic coefficient curves}
    \end{subfigure}
    \begin{subfigure}{1\textwidth}
        \centering
        \includegraphics[width=0.4\linewidth]{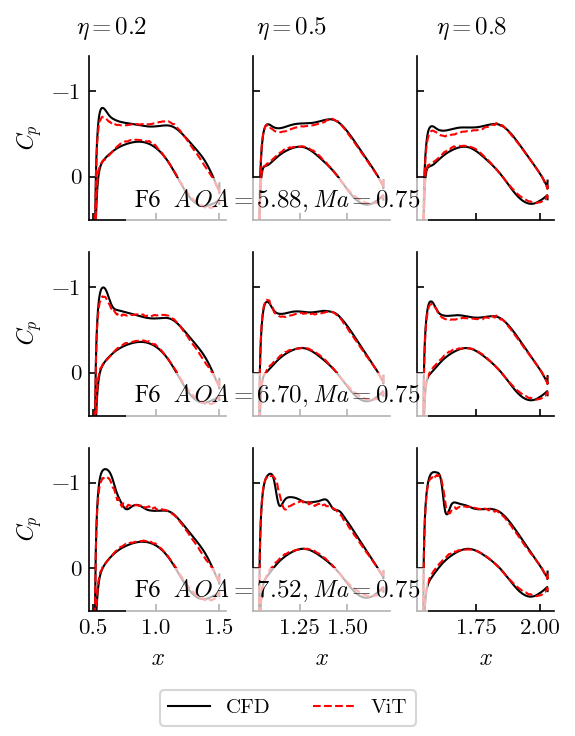}
        \includegraphics[width=0.42\linewidth]{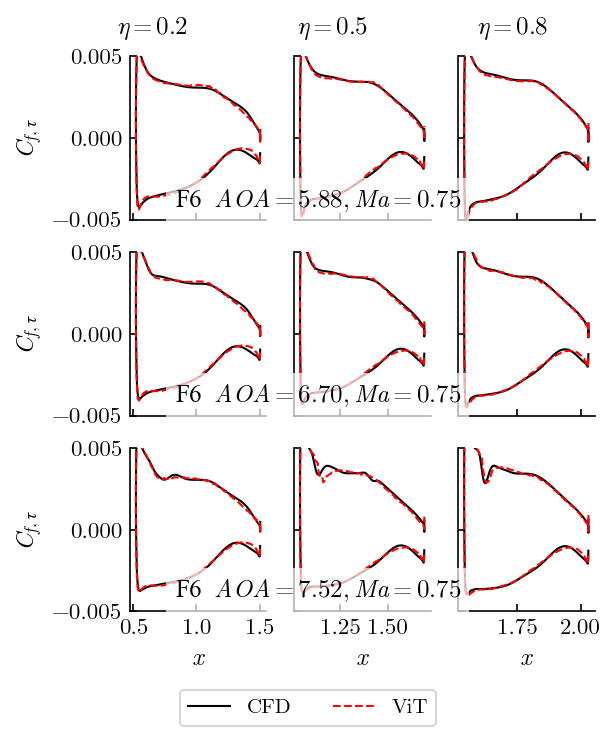}
        \caption{Cross-sectional coefficients distribution}
    \end{subfigure}
        
    \caption{Surface flow and aerodynamic coefficients prediction of the DLR-F6 wing}
    \label{fig:testbenchf6}
\end{figure}

\begin{figure}[H]
    \centering
    \begin{subfigure}{0.495\textwidth}
        \centering
        \includegraphics[width=\linewidth]{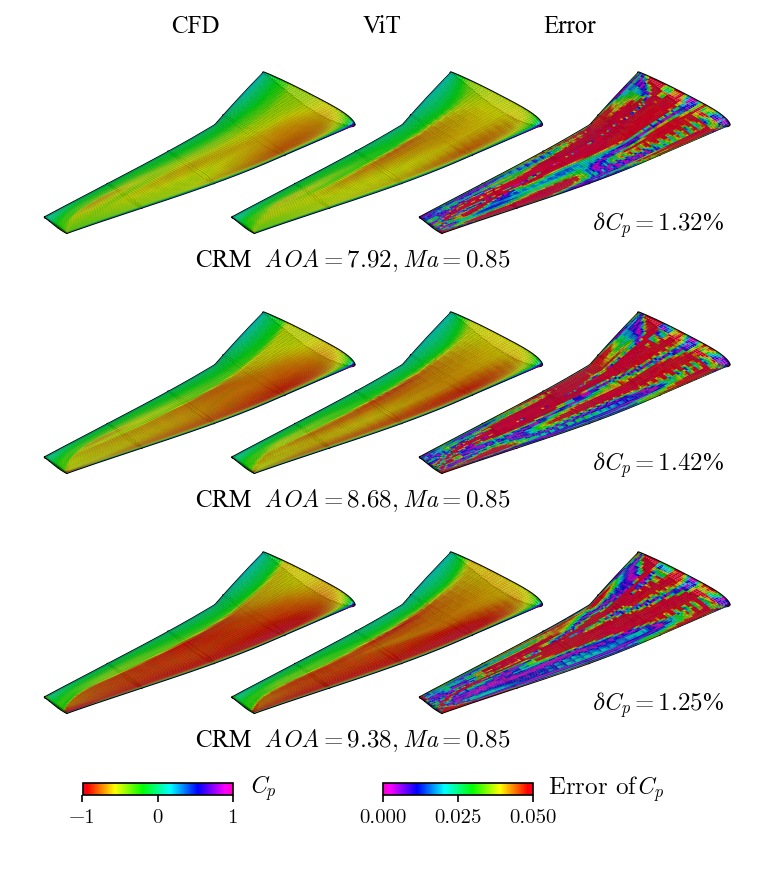}
        \caption{Surface contours}
    \end{subfigure}
    \begin{subfigure}{0.4\textwidth}
        \centering
        \includegraphics[width=\linewidth]{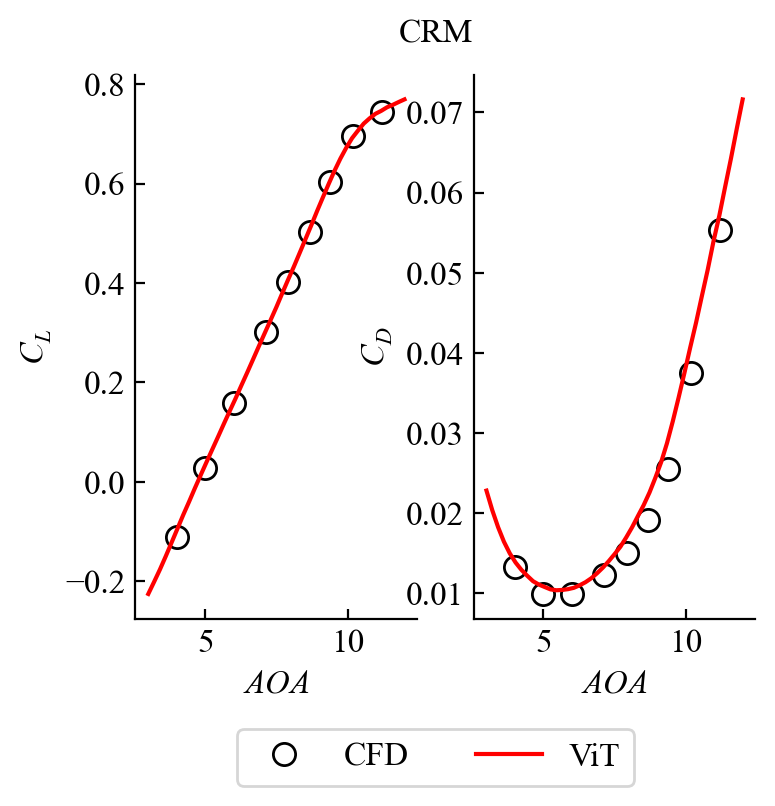}
        \caption{Aerodynamic coefficient curves}
    \end{subfigure}
    \begin{subfigure}{1\textwidth}
        \centering
        \includegraphics[width=0.4\linewidth]{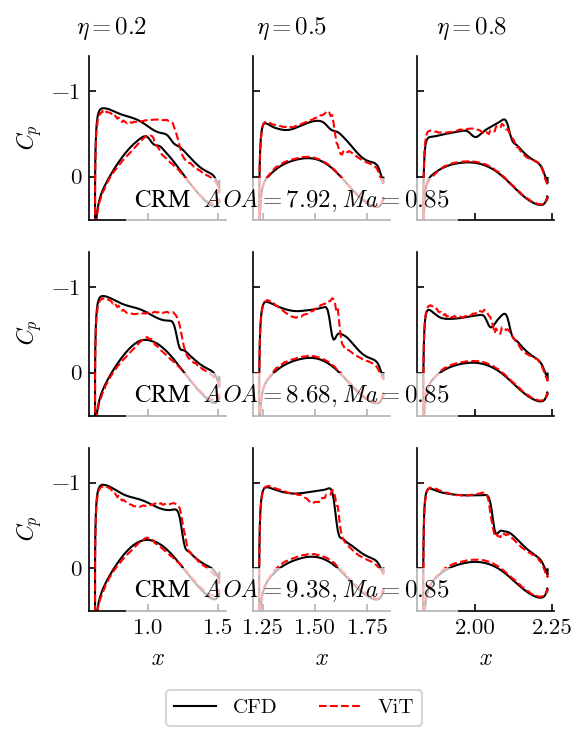}
        \includegraphics[width=0.42\linewidth]{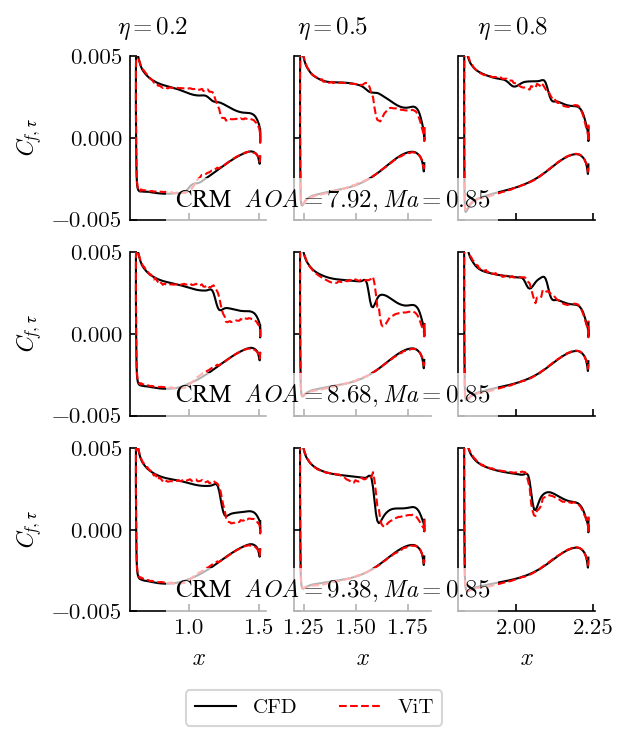}
        \caption{Cross-sectional coefficients distribution}
    \end{subfigure}
        
    \caption{Surface flow and aerodynamic coefficients prediction of the CRM wing}
    \label{fig:testbenchcrm}
\end{figure}
\end{modifiedblock}

\bibliographystyle{unsrt}
\bibliography{sample}

\appendix

\end{document}